\documentclass{article}
\usepackage{iclr2027_conference,times}
\usepackage[T1]{fontenc}
\usepackage[utf8]{inputenc}
\usepackage{amsmath,amssymb,amsthm,mathtools}
\usepackage{graphicx,booktabs,tabularx,array,multirow,wrapfig,pifont}
\usepackage[caption=false,font=footnotesize]{subfig}
\usepackage{xcolor,tikz,pgfplots,microtype,url}
\pgfplotsset{compat=1.18}
\usetikzlibrary{arrows.meta,positioning,calc}
\usepackage[most]{tcolorbox}
\usepackage[hidelinks]{hyperref}
\newcommand{\E}{\mathbb{E}}

\newcommand{\I}{\mathcal{I}}
\newcommand{\Sset}{\mathcal{S}}
\newcommand{\Aset}{\mathcal{A}}
\newcommand{\KL}{\mathrm{KL}}

\newcommand{\sm}{\operatorname{softmax}}
\newcommand{\diag}{\operatorname{diag}}

\newcommand{\one}{\mathbf{1}}
\newcommand{\proj}{P_{\!\perp}}
\newcommand{\hF}{\widehat F}
\newcommand{\bF}{\bar F}
\newcommand{\norm}[1]{\left\lVert#1\right\rVert}

\newcommand{\inner}[2]{\left\langle#1,#2\right\rangle}
\newcommand{\piref}{\pi^{\mathrm{ref}}}
\newcommand{\pistar}{\Pi^*}
\definecolor{pendingblue}{RGB}{0,70,180}

\definecolor{takeawayaccent}{RGB}{26,107,124}
\newtcolorbox{takeawaybox}[1][Takeaway]{enhanced,arc=2pt,outer arc=2pt,boxrule=0.6pt,colframe=takeawayaccent!55,colback=takeawayaccent!5,boxsep=0pt,left=7pt,right=7pt,top=3pt,bottom=3pt,before skip=6pt,after skip=6pt,fontupper=\normalsize,attach boxed title to top left={xshift=7pt,yshift*=-\tcboxedtitleheight/2},boxed title style={boxrule=0pt,arc=1.5pt,colback=takeawayaccent,colframe=takeawayaccent,boxsep=0pt,left=4pt,right=4pt,top=1.5pt,bottom=1.5pt},fonttitle=\small\bfseries\color{white},title={#1}}
\newcommand{\takeaway}[2][Takeaway]{\begin{takeawaybox}[#1]#2\end{takeawaybox}}
\newtheorem{theorem}{Theorem}[section]
\newtheorem{proposition}[theorem]{Proposition}
\newtheorem{corollary}[theorem]{Corollary}
\newtheorem{lemma}[theorem]{Lemma}
\theoremstyle{definition}

\theoremstyle{remark}

\newcolumntype{Y}{>{\raggedright\arraybackslash}X}
\AddToHook{env/table/begin}{\setlength{\abovecaptionskip}{0pt}\setlength{\belowcaptionskip}{4pt}}

\newcommand{\rb}[1]{} 

\hypersetup{pdftitle={Your Policy Is Secretly a Harness: The Decision Geometry of Agent Interfaces},pdfauthor={Jian Mu, Ruibo Shen, Yao Shu}}
\title{Your Policy Is Secretly a Harness:\\The Decision Geometry of Agent Interfaces}
\author{\parbox[t]{\dimexpr\textwidth-2\tabcolsep\relax}{\normalfont\centering
\begin{tabular}[t]{@{}c@{}}
\textbf{Jian Mu}\\
The Hong Kong University of\\
Science and Technology (Guangzhou)
\end{tabular}\hspace{0.18\textwidth}%
\begin{tabular}[t]{@{}c@{}}
\textbf{Ruibo Shen}\\
The Hong Kong University of\\
Science and Technology (Guangzhou)
\end{tabular}\\[10pt]
\begin{tabular}[t]{@{}c@{}}
\textbf{Yao Shu}\thanks{Corresponding author: \texttt{yaoshu@hkust-gz.edu.cn}.}\\
The Hong Kong University of Science and Technology (Guangzhou)\\
The Hong Kong University of Science and Technology
\end{tabular}}}
\iclrfinalcopy 
\begin{document}
\maketitle
\pagestyle{plain} 

\setlength{\textfloatsep}{12pt plus 2pt minus 2pt}
\setlength{\abovecaptionskip}{4pt}
\begin{abstract}
Language model agents are post-trained and deployed through a harness, the prompts, tools, and control logic that present a task to the model and execute its outputs. Harnesses are often revised and judged by how an already trained model performs under them, which assumes that what the model learned under one harness carries over to the next. We ask what a policy learns from its harness and what survives when the harness changes. Under KL-regularized control, we show that the log odds of an optimal policy recover how strongly its harness favors each action over the others, so a trained policy carries the preferences of its harness. Deployed return then equals the ceiling of the harness, the best return it offers, minus the mismatch of the policy, the return lost to its own suboptimal decisions. Training changes only the mismatch, while changing the harness moves both terms. Comparing harnesses with a fixed model therefore mixes what each harness offers with how well that model uses it. In MiniWoB++, post-training makes a model's preferences specific to its training harness. Reversing the order of the observation fields leaves every optimal decision unchanged yet lowers a trained model's success from 85.0\% to 19.2\%. Conversely, every new harness that a fixed-model comparison rejects but that raises the ceiling outperforms the original agent after further training under it, and ceiling and mismatch measured beforehand predict these gains. For coding-agent harnesses on Terminal-Bench Lite, ceiling minus mismatch estimated with a stronger model as a stand-in for the optimum reproduces the accuracy ranking, while either term alone does not.
Code is available at \url{https://github.com/2020-qqtcg/HRP}.
\end{abstract}

\section{Introduction}
\label{sec:intro}
Language model agents are post-trained and deployed through a harness, the prompts, tools, and control logic that present a task to the model and execute its outputs. Interface design and prompt formatting change agent performance \citep{yang2024sweagent,sclar2024format}. Automated harness design revises prompts, tools, and workflows and scores each candidate by how a fixed model performs under it \citep{hu2025adas,zhang2025aflow,fernando2023promptbreederselfreferentialselfimprovementprompt,zhang2026darwingodelmachineopenended}. Such a score assumes that what the model learned under earlier harnesses carries over to each candidate, yet the same interface change can raise one model's accuracy and lower that of a fine-tuned model \citep{mu2025gui360}. At the same time, training can move behavior that a prompt or a harness supplies into the weights \citep{snell2022context,ye2026harnesszero}, and recent methods adapt harnesses and weights together \citep{hebbar2026sia,chen2026coharness,kim2026whale}. These practices raise the question we study:

\emph{What does a policy learn from its harness, and what survives when the harness changes?}

Two experiments make the question concrete. In the first, we take a web agent post-trained under one harness and reverse the order of the fields in its observation, a JSON object with fields such as the task instruction, the page state, and the available actions. The fields keep their contents. We also hold fixed the actions, their execution, the rewards, and the reference policy that regularizes training, so every optimal decision stays the same. The model still changes its most probable action at about half of the states, and its task success falls from 85.0\% to 19.2\%. Keeping the decisions a model should make does not keep the decisions it makes.

\begin{wrapfigure}{l}{0.48\linewidth}
\centering
\vspace{-12pt}
\setlength{\abovecaptionskip}{3pt}
\includegraphics[width=0.9\linewidth]{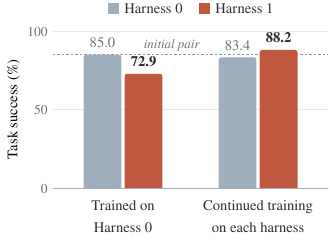}
\caption{Adaptation reverses a frozen comparison. The left group evaluates the model trained under Harness~0, the source checkpoint, under both harnesses with its weights frozen. The right group evaluates the models after equal further training under each harness. The dotted line marks the source checkpoint's success under Harness~0, and colors mark the evaluation harness.}
\label{fig:intro-success}
\vspace{-10pt}
\end{wrapfigure}
In the second, a new harness adds batched clicks, which execute two clicks in one decision. The new actions raise the best achievable return, yet the trained model's success drops when we move it to the new harness without changing its weights. A frozen comparison, which evaluates candidate harnesses with the same trained weights, would therefore reject the new harness. We then train the model further for the same budget under each harness, starting from the same trained weights, which we call the source checkpoint. We call further training under the new harness adaptation, and further training under the original harness continuation. The adapted model succeeds more often than both the source checkpoint and the continued one. The harness that first hurt the model gives the best agent. Figure~\ref{fig:intro-success} shows the task success at each step, with the new harness as Harness~1.

We explain both results in the control problem that a harness induces over semantic actions, the operations an agent executes together with their arguments. Under KL-regularized control, the optimal policy is a softmax over action scores. Each score adds an action's long-term value to its log probability under the reference policy, so an optimal policy's log odds between two actions are proportional to the difference of their scores. A policy trained to optimality under a harness therefore carries the relative action scores that the harness induces, even though it contains none of the harness's code, tools, or execution machinery. In this sense, a policy is secretly a harness. In our controlled study, real post-training moves a model's action scores toward the optimum of its training harness, and each harness leaves its own preferences in the model.

\begin{wrapfigure}[25]{r}{0.46\linewidth}
\centering
\vspace{-12pt}
\setlength{\abovecaptionskip}{3pt}
\includegraphics[width=\linewidth,trim=0 34 0 16,clip]{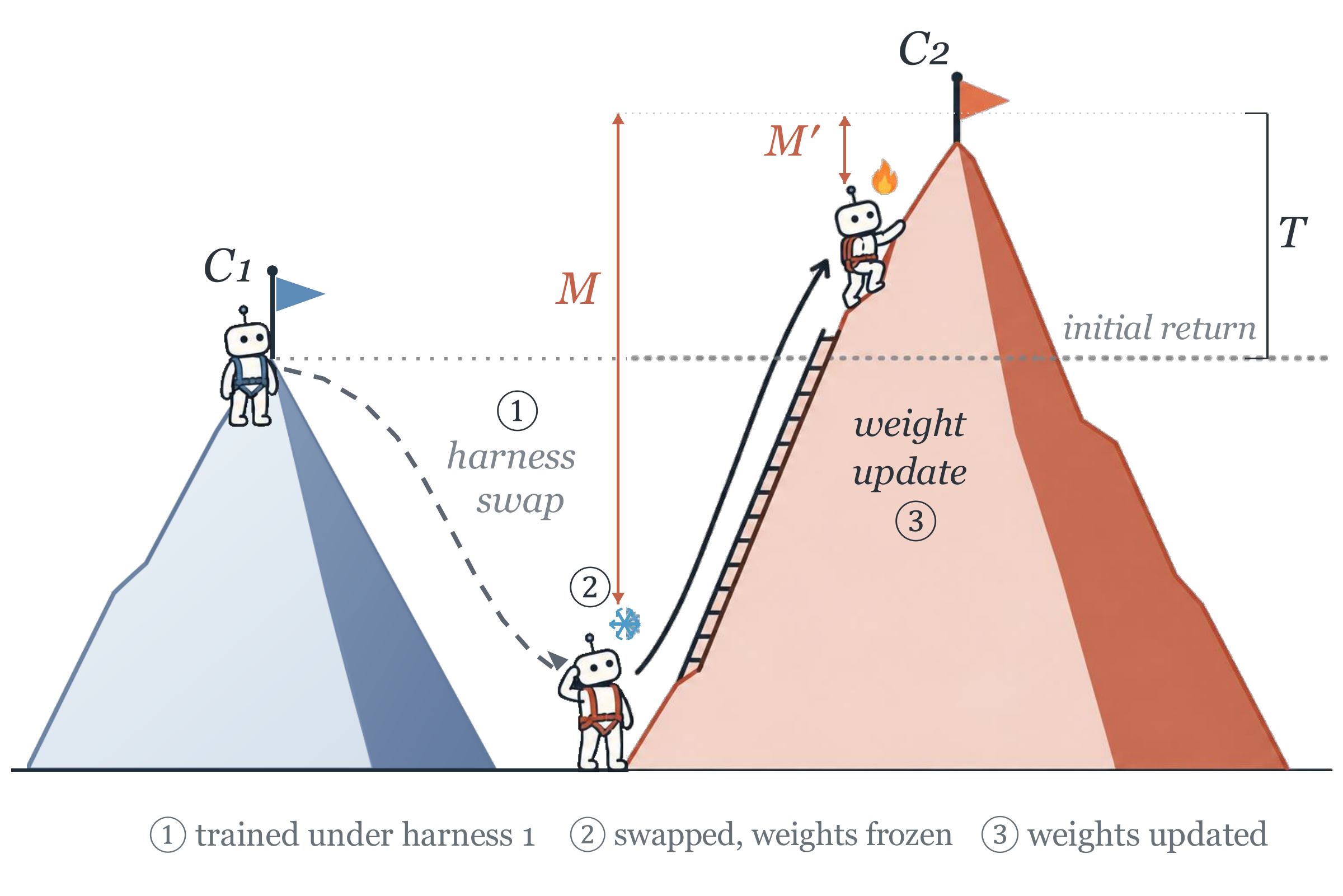}
\caption{A harness revision and a weight update act on different terms. Summit heights are the ceilings, the robot's altitude is its return, and the gap between them is the mismatch. Moving the trained model to the new harness with frozen weights (\protect\ding{172} to \protect\ding{173}) raises the ceiling but lowers the return. Updating the weights under the new harness (\protect\ding{173} to \protect\ding{174}) lowers only the mismatch. The adapted model beats the source checkpoint under the original harness once its mismatch $M'$ falls below $T$, the new ceiling minus the source checkpoint's return under the original harness.}
\label{fig:overview}
\vspace{-10pt}
\end{wrapfigure}
Action scores also show why a trained policy can fail after it is moved to a new harness. The scores a policy carries are the ones it expresses through its training interface, which we call its readout through that interface, and the same weights can express different scores through another interface. A policy is thus secretly its training harness, as read through that harness. A new harness can therefore change which actions are optimal, by moving the optimal scores, and separately change which actions the model picks, by moving the readout. Reversing the observation fields leaves the optimal scores in place and moves only the readout.

Scores fix decisions but not how much return a harness offers, since raising every reward by the same constant changes no decision. To convert score changes into return, we therefore add this value back by applying the regularized performance-gap identity \citep{geist2019regularized,odonoghue2022bregman} to the harness-induced problem. Deployed return equals the harness's ceiling $C$, the best return any policy can reach under it, minus the policy's mismatch $M$, the return the policy loses by departing from the optimal policy. The mismatch is the KL divergence from the optimal policy, accumulated along the policy's own trajectories. The ceiling does not depend on the model's weights, so training under a fixed harness changes only $M$, while a harness change can move both terms. A frozen comparison of harnesses therefore mixes what each harness offers with how well the current checkpoint uses it. In the batched-click example, the new harness raises the ceiling slightly but raises the frozen model's mismatch about ten times as much, so return falls. Adaptation then lowers the mismatch, and once the mismatch falls below a threshold we derive, the adapted model beats the original model under the original harness. Figure~\ref{fig:overview} illustrates both steps. Our contributions are as follows.

\noindent\textbf{1. What a policy learns under a harness.}
The log odds of an optimal policy identify exactly the relative action scores that a harness induces. These scores fix every decision but not how much return the harness offers, which Contribution~3 adds back as the ceiling. On MiniWoB++, post-training moves the model's action preferences toward the training harness's optimum and makes them specific to that harness.

\noindent\textbf{2. Why transfer fails.}
We split the score error of a policy moved to a new harness into three parts. The target shift is the change in the optimal scores, the training residual is the error that training under the original harness left behind, and the readout drift is the change in the scores that the model expresses. Presentation changes that keep every optimal decision isolate the readout drift, and they alone can make a trained model fail on most tasks.

\noindent\textbf{3. What transfer costs and how to measure it.}
Deployed return equals ceiling minus mismatch exactly. From this identity we derive the mismatch threshold below which adaptation turns a harness revision that hurt the frozen model into a gain. Adaptation to batched clicks crosses this threshold and outperforms continued training under the original harness. We also move trained checkpoints between six harnesses. Among the moves that a frozen comparison rejects, ceiling and mismatch measured before adaptation predict which ones adaptation makes beneficial. Mismatch estimated from sampled actions predicts return changes when the optimum is known. For coding agents, where no optimum is available, estimates that put a stronger teacher model in the optimum's place order five harnesses by Terminal-Bench Lite accuracy.

\section{Harnesses Induce Semantic Control Problems}
\label{sec:setup}
A harness turns model outputs into executed operations. We call an executed operation together with its arguments a semantic action $a\in\Aset$, such as clicking a particular element or running a query. Different outputs can execute the same action. For example, a JSON object and a function-call string may issue the same parameterized query. Let $s\in\Sset$ denote the semantic decision state. An interface $\I=(c_\I,d_\I,e_\I)$ formalizes a harness. It encodes the state into a context, decodes a model response into an action, and executes that action. For a token policy $\pi_\theta$, the induced semantic policy sums the probabilities of the outputs that decode to the same action:
\begin{equation}
\Pi_{\theta,\I}(a\mid s)
=\sum_{y:\,d_\I(y;s)=a}\pi_\theta(y\mid c_\I(s)).
\label{eq:pushforward}
\end{equation}
Both the context $c_\I$ and the decoder $d_\I$ enter Eq.~\eqref{eq:pushforward}, so fixed weights do not fix the semantic policy.

Together with a fixed task environment, $\I$ induces a transition law $P_\I$, a utility $r_\I$ that includes execution costs, and a semantic reference policy $\piref_\I$. Every call to the evaluated model is a decision, including calls for verification or retry, and the state includes the harness's controller memory, such as tool results and remaining budget. Appendix~\ref{app:semantics} specifies this construction and the conditions under which training over tokens yields semantic optimality.

Across comparisons, we fix the initial distribution $\rho_0$, discount $0\le\gamma<1$, and regularization coefficient $\tau>0$, and measure rewards on a common scale. Each interface's reference is part of its control problem and does not change during training. For finite $\Sset,\Aset$, bounded rewards, and $\piref_\I(a\mid s)>0$, the regularized return of a semantic policy $\Pi$ is
\begin{equation}
J_\I(\Pi)=\E_{\rho_0,\Pi,P_\I}\!\left[\sum_{t\ge0}\gamma^t
\left(r_\I(s_t,a_t)-\tau\log\frac{\Pi(a_t\mid s_t)}{\piref_\I(a_t\mid s_t)}\right)\right].
\label{eq:objective}
\end{equation}
Write $\pistar_\I$ for the optimal policy and $d^\Pi_\I(s)=(1-\gamma)\sum_{t\ge0}\gamma^t\Pr_\I^\Pi(s_t=s)$ for the discounted occupancy of $\Pi$.

\paragraph{Controlled study.}
\label{sec:exp-setting}
In our mechanism experiments, every quantity above can be computed exactly. Six harnesses act on three MiniWoB++ task families~\citep{liu2018reinforcement}, and we compile each induced control problem into a finite graph. H0 executes one native action per decision, and H1--H5 add batched clicks, automatic checks, verification tools, or extra instructions. Each prompt lists the complete action menu with one token per action, so the model's next-token distribution gives $\Pi_{\theta,\I}$ exactly. The reward is binary task success. We post-train Qwen3.5-2B with exact policy gradients, and the frozen base model serves as the reference. A soft Bellman solve gives each harness's optimum. Evaluations use tasks held out from training, and Appendix~\ref{app:e1} gives the full protocol.

\section{What a Policy Learns under a Harness}
\label{sec:framework}
We now ask what a model's action probabilities reveal about the control problem it was trained on.

\subsection{Optimal Policies Encode Relative Action Scores}
\label{sec:identification}
The optimal action value $Q^*_\I$ depends on the harness's utility and transition law through the soft Bellman equations, and the optimal policy takes the Gibbs form $\pistar_\I(a\mid s)\propto\piref_\I(a\mid s)\exp(Q^*_\I(s,a)/\tau)$ \citep{haarnoja2017reinforcementlearningdeepenergybased,geist2019regularized}. An action's optimal probability therefore depends on its long-term value and its reference probability. We combine the two into the optimal action score
\begin{equation}
F^*_{\I}(s,a)=Q^*_{\I}(s,a)+\tau\log\piref_{\I}(a\mid s),\qquad
\pistar_\I(\cdot\mid s)=\sm\bigl(F^*_\I(s,\cdot)/\tau\bigr).
\label{eq:interface-energy}
\end{equation}
For instance, an optimal policy may favor a batch of clicks because the batch leads to better outcomes or because the reference places more weight on it. The score records the combined effect.

Adding the same constant to every action's score at a state leaves the policy unchanged, so only score differences within a state affect decisions. We write $[F]$ for the class of score fields related by such statewise constants. Two interfaces have \emph{equivalent optimal decisions} when $\pistar_{\I_1}(\cdot\mid s)=\pistar_{\I_2}(\cdot\mid s)$ at every aligned state, that is, every semantic state that the two interfaces share.

\begin{proposition}[Behavioral identification]
\label{thm:identification}
Two interfaces have equivalent optimal decisions if and only if $[F^*_{\I_1}]=[F^*_{\I_2}]$. If a checkpoint is optimal for its training interface at every semantic state, then
\begin{equation}
[\tau\log\Pi_{\theta,\I_{\rm tr}}]=[F^*_{\I_{\rm tr}}],\qquad
\tau\log\frac{\Pi_{\theta,\I_{\rm tr}}(a\mid s)}{\Pi_{\theta,\I_{\rm tr}}(b\mid s)}
=F^*_{\I_{\rm tr}}(s,a)-F^*_{\I_{\rm tr}}(s,b).
\label{eq:identification}
\end{equation}
\end{proposition}
Behavior determines exactly these relative scores. Different reference policies, rewards, and even transition laws can share one optimal score field and therefore one optimal policy, in line with partial identifiability results for reward learning \citep{skalse2023invariancepolicyoptimisationpartial}. Behavior also leaves absolute value undetermined, since adding a constant to every reward raises the optimal return and changes no decision. A policy can thus carry the decisions a harness induces without its code, tools, or execution machinery. The value a harness offers is a separate quantity, which Section~\ref{sec:return} adds back. Appendix~\ref{app:identification} proves the proposition from the Gibbs form and constructs such mechanisms.

A real checkpoint departs from the optimum. To compare the two, we represent each class by its centered member, subtracting the mean score across actions at each state: $\bF=\proj F$ with $\proj=I-\one\one^\top/n$ and $n=|\Aset|$. For positive semantic probabilities, the checkpoint's centered scores are $\hF_{\theta,\I}=\tau\proj\log\Pi_{\theta,\I}$, the scores whose softmax in Eq.~\eqref{eq:interface-energy} reproduces the checkpoint's own policy. We call them the model's readout through $\I$, the action preferences it expresses through the interface. Its score error under the training interface, $e_{\rm tr}=\hF_{\theta,\I_{\rm tr}}-\bF^*_{\I_{\rm tr}}$, vanishes exactly when the checkpoint is optimal.

\subsection{Post-Training Makes Preferences Harness-Specific}
\label{sec:exp-training}
Proposition~\ref{thm:identification} concerns optimal policies. We ask whether real post-training moves a model toward the optimum of its training harness, and whether what it learns depends on that harness. We post-train Qwen3.5-2B under each of H0--H5. The geometry error of a checkpoint is the mean squared difference between its centered scores and the optimal ones, averaged over a fixed bank of reachable states, so that every checkpoint is scored on the same decisions.

\begin{figure}[t]
\centering
\includegraphics[width=0.9\linewidth,trim=0 6 0 0,clip]{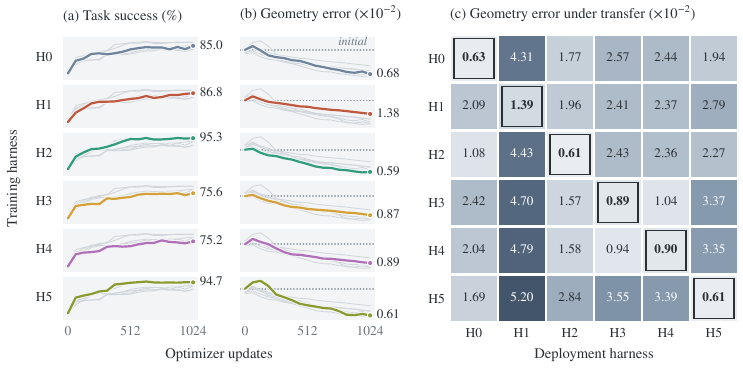}
\caption{Post-training moves action preferences toward the optimum of the training harness and makes them specific to it. Rows denote training harnesses. (a) Task success and (b) geometry error relative to the training harness's optimum during post-training. Colored curves show the harness of each row, gray curves the other harnesses, and dotted lines the error at initialization. (c) Geometry error of each final checkpoint under each deployment harness. Outlined entries mark the lowest error in each column, attained by the checkpoint trained under that harness.}
\vspace{-4px}
\label{fig:training-geometry}
\end{figure}

Under every harness, post-training raises task success, and the geometry error falls well below its initial level, by 69\% to 90\% by the end of training. It falls as much on held-out tasks that play no role in the updates as on training tasks. The model learns effective task behavior while its action scores approach the optimum of its training harness.

To test whether the learned preferences depend on the training harness, we deploy each final checkpoint under all six harnesses on test tasks. A fixed deployment harness fixes the dynamics, reference, and optimal scores, so the six checkpoints are compared against the same target. Under every deployment harness, the checkpoint trained under that harness has the lowest error. Post-training thus makes the model's preferences specific to its training harness. Figure~\ref{fig:training-geometry} shows the learning curves in its first two panels and this comparison in the third.

\takeaway[Takeaways: What a policy learns]{A policy trained to optimality under a harness carries that harness's relative action scores. Real post-training moves a model's action scores toward these optimal scores and makes them specific to the training harness.}

\section{Why Policy Transfer Fails}
\label{sec:transfer}
Section~\ref{sec:framework} shows that a trained policy carries the relative action scores of its training harness. It carries them in its readout through the training interface rather than independently of that interface, so the same weights need not express the same preferences under another harness. A harness change can therefore break a policy that has learned its training harness well.

\subsection{A Harness Change Moves the Target and the Readout}
\label{sec:swap}
Let interfaces $1$ and $2$ be the training and deployment interfaces. A change in execution can change which actions are optimal, and a change in presentation can change which actions the frozen model selects. On aligned states and actions, the deployed score error splits into three terms:
\begin{equation}
\underbrace{\hF_{\theta,2}-\bF^*_2}_{\text{deployed error}}
=\underbrace{\bF^*_1-\bF^*_2}_{g_{12}:\ \text{target shift}}
+\underbrace{\hF_{\theta,1}-\bF^*_1}_{e_1:\ \text{training residual}}
+\underbrace{\hF_{\theta,2}-\hF_{\theta,1}}_{\eta_{12}:\ \text{readout drift}}.
\label{eq:three-errors}
\end{equation}
The training residual exists before the change, and the target shift and the readout drift are its two effects. They are vectors, so they can add up or cancel.

\subsection{Equivalent Optimal Decisions, Different Behavior}
\label{sec:presentation}
\begin{wraptable}[14]{r}{0.47\linewidth}
\vspace{-\intextsep}
\centering\small
\setlength{\abovecaptionskip}{0pt}
\setlength{\belowcaptionskip}{4pt}
\caption{Equivalent optimal decisions, different choices. Choice change is the share of states whose most probable action changes from the original. $M$ and $J$ are the mismatch and regularized return of Section~\ref{sec:ceiling}.}
\label{tab:readout}
\setlength{\tabcolsep}{3pt}
\renewcommand{\arraystretch}{1.1}
\begin{tabular}{@{}lrrrr@{}}
\toprule
Presentation & \shortstack{Choice\\change} & $M$ & $J$ & \shortstack{Task\\success}\\
\midrule
Original & 0.0\% & 0.015 & $0.423$ & 85.0\%\\
Reversed fields & 52.0\% & 0.484 & $-0.046$ & 19.2\%\\
Menu as list & 9.2\% & 0.023 & $0.415$ & 86.0\%\\
Shifted codes & 27.7\% & 0.148 & $0.289$ & 57.1\%\\
Reversed codes & 30.0\% & 0.130 & $0.308$ & 68.2\%\\
Permuted codes & 25.3\% & 0.110 & $0.328$ & 67.6\%\\
\bottomrule
\end{tabular}
\end{wraptable}
When two interfaces have equivalent optimal decisions, $g_{12}=0$, and any change in the model's error comes from readout drift. We test this with the H0-trained checkpoint. We reverse the order of the observation fields, render the action menu as a list, or reassign the action codes, and keep the transitions, rewards, and semantic reference fixed, so neither the target nor the reference can move. The optimal policy and the optimal return thus stay the same, as numerical checks confirm.

Equivalent optimal decisions do not produce equivalent behavior. Reversing the field order has the largest effect. It changes the model's most probable action at about half of the states, and most tasks then fail. Reassigning the action codes has a moderate effect, and the list format has a small one. Unlike earlier studies of prompt-format sensitivity \citep{sclar2024format}, the control problem here certifies that the optimal policy is unchanged, so the entire effect comes from the model's readout. Table~\ref{tab:readout} lists the results, Appendix~\ref{app:exp-readout} gives the protocol, and Appendix~\ref{app:correspondence} bounds readout drift by the model's sensitivity to corresponding outputs.

\takeaway[Takeaways: Why transfer fails]{A harness change can move the optimal decisions and the model's choices separately. A new presentation alone can make a trained model fail on most tasks.}

\section{What Transfer Costs and When Adaptation Recovers It}
\label{sec:return}
Separating the two changes shows what moved. What they cost in return depends on which decisions they affect and how often the policy reaches them.

\subsection{From Score Errors to Lost Return}
\label{sec:decomposition}
Consider a single decision at state $s$ in a fixed interface $\I$. If the policy $\Pi$ makes this decision and the optimal policy makes every later one, the return lost relative to acting optimally throughout is $\tau\KL(\Pi(\cdot\mid s)\|\pistar_\I(\cdot\mid s))$. If $\Pi$ also makes the later decisions, each of them adds its own loss, and summing along the trajectories of $\Pi$ gives the total gap \citep{odonoghue2022bregman}.
\begin{proposition}[Regularized performance gap]
\label{thm:gap}
For any semantic policy $\Pi$ in a fixed interface,
\begin{equation}
J_\I(\pistar_\I)-J_\I(\Pi)=\frac{\tau}{1-\gamma}\E_{s\sim d^\Pi_\I}\KL\bigl(\Pi(\cdot\mid s)\|\pistar_\I(\cdot\mid s)\bigr).
\label{eq:gap}
\end{equation}
\end{proposition}
The occupancy weights each local loss by how often the evaluated policy reaches the state, so errors on the policy's common paths cost the most. In terms of scores, the local loss is the Bregman divergence $D_\Phi(F,G)$ of $\Phi(F)=\tau\log\sum_a e^{F_a/\tau}$, a smooth maximum of the scores. $D_\Phi(F,G)$ is the gap between $\Phi(F)$ and the linear approximation of $\Phi$ at $G$. For positive semantic action probabilities,
\begin{equation}
\tau\KL(\Pi_{\theta,\I}\|\pistar_\I)=D_\Phi(F^*_\I,\hF_{\theta,\I}).
\label{eq:local-energy-loss}
\end{equation}
The scores of the actions at a state locate the model's decision, and $D_\Phi$ measures how much return the decision loses when the model's scores sit at $\hF_{\theta,\I}$ instead of $F^*_\I$. We call these scores and this cost together the decision geometry of a harness. Averaging $D_\Phi$ under the deployed policy's occupancy turns the score error of Eq.~\eqref{eq:three-errors} into lost return. For small errors, $D_\Phi$ grows roughly quadratically with the summed error, so the three terms of Eq.~\eqref{eq:three-errors} cost return through their sum rather than one by one, and Proposition~\ref{thm:probe-gap} below turns this into a bound. Appendix~\ref{app:geometry} proves both identities.

\subsection{Ceiling and Mismatch}
\label{sec:ceiling}
Relative scores fix the decisions and leave the value of a harness undetermined. We add that value back. Define the ceiling $C(\I)=J_\I(\pistar_\I)$, the optimal regularized return of the harness and thus the best that any policy can reach under it, and the mismatch $M(\theta,\I)$, the return the checkpoint leaves unrealized. Proposition~\ref{thm:gap} gives
\begin{equation}
J_\I(\Pi_{\theta,\I})=C(\I)-M(\theta,\I),\qquad
M(\theta,\I)=\frac{\tau}{1-\gamma}\E_{d^{\Pi_{\theta,\I}}_\I}\KL(\Pi_{\theta,\I}\|\pistar_\I).
\label{eq:ceiling-mismatch}
\end{equation}
The ceiling depends on the harness and the objective, and the mismatch depends on the harness and the checkpoint, including errors inherited from training. With the harness fixed, a weight update changes return only through $M$, and a harness change can move both terms. In the summit illustration of Figure~\ref{fig:overview}, the step from \ding{172} to \ding{173} is a harness revision, and the step from \ding{173} to \ding{174} is a weight update. We call a checkpoint together with the harness it is deployed under a pair. For a frozen checkpoint, write $J_j$, $C_j$, and $M_j$ for the quantities under interface $j$. Then
\begin{equation}
J_2-J_1=(C_2-C_1)-(M_2-M_1),
\label{eq:reversal}
\end{equation}
so a revision helps a frozen model only when its ceiling gain exceeds the added mismatch. In the presentation study, the ceiling is fixed, so each increase in mismatch appears as an equal loss in return. Table~\ref{tab:readout} reports both for each presentation.

\subsection{When Adaptation Makes a Revision Beneficial}
\label{sec:adaptation}
When a revision raises the ceiling but lowers the frozen model's return, adapting to it can realize the gain, since every unit of mismatch removed adds a unit of return. Write $J(\theta,\I)=J_\I(\Pi_{\theta,\I})$.
\begin{corollary}[Adaptation after a harmful revision]
\label{cor:adaptation}
Let $\theta_0$ be the source checkpoint, trained under $\I_1$, let $\theta_1$ be the checkpoint adapted from it under $\I_2$, and let $\Delta C=C_2-C_1$. Suppose the revision harms the frozen source checkpoint, $J(\theta_0,\I_2)<J(\theta_0,\I_1)$. The adapted pair outperforms the source pair, $J(\theta_1,\I_2)>J(\theta_0,\I_1)$, if and only if
\begin{equation}
M(\theta_1,\I_2)<C_2-J(\theta_0,\I_1)=\Delta C+M(\theta_0,\I_1).
\label{eq:adaptation-threshold}
\end{equation}
\end{corollary}
The threshold $T=\Delta C+M(\theta_0,\I_1)$ is the new ceiling minus the source pair's return. In words, adaptation pays off once the mismatch left under the new harness is smaller than the ceiling gain plus the mismatch that the source pair already had. Three comparisons describe the outcome. \emph{Recovery} compares the adapted pair with the source pair, \emph{harness ranking} compares the two harnesses at the adapted checkpoint, and \emph{gain over continuation} compares adaptation with continued training under the source harness at the same budget, which tests whether the benefit comes from the revision or from further training. Appendix~\ref{app:adaptation} proves the corollary, and Appendix~\ref{app:bandit-reversal} illustrates reversal and adaptation in a two-action example.

\begin{figure}[t]
\centering
\includegraphics[width=\linewidth]{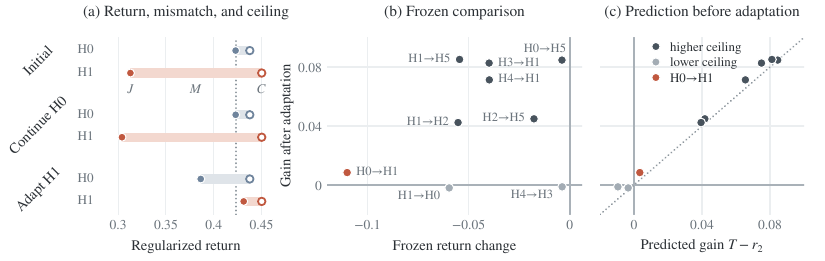}
\caption{Harmful revisions become beneficial after adaptation when they raise the ceiling. (a) Return $J$, ceiling $C$, and mismatch $M$ of the H0-trained checkpoint, its H0 continuation, and its H1 adaptation under each harness. The dotted line marks the source pair's return. Gains of the adapted pair over the source pair in the nine swaps that a frozen comparison rejects and that have $T>0$, plotted against (b) the frozen return change and (c) the predicted gain $T-r_2$. In (b) and (c), dark points raise the ceiling, light points lower it, and the red point is H0$\to$H1.}
\vspace{-4px}
\label{fig:adaptation}
\end{figure}

\paragraph{Batched clicks.}
\label{sec:exp-adaptation}
From the H0-trained checkpoint, we create two branches with identical weights, update budgets, and task draws. One continues training under H0, and the other adapts to H1, which adds batched clicks. H1 raises the ceiling slightly, but the frozen checkpoint carries its H0 preferences into H1 and adds far more mismatch, so return and task success fall. Adaptation under H1 removes most of this mismatch, while the H0 branch, which never trains under H1, keeps it. During adaptation, the checkpoint's return under H1 rises and its return under H0 falls. Its ranking of the two harnesses therefore reverses first, and only later does it beat the source pair. At the end, the adapted pair beats both the source pair and the H0 branch in task success and in return, and the paired intervals of these differences exclude zero. Figure~\ref{fig:adaptation} shows both branches in its first panel, and Appendix~\ref{app:exp-adaptation} gives the endpoint measurements.

\paragraph{Across harness swaps.}
Batched clicks are one of many possible swaps among H0--H5. A swap moves the checkpoint post-trained under a source harness to a different target harness. We adapt every swap that a frozen comparison rejects and whose $T$ is positive. When $T\le0$, no policy under the target harness can beat the source pair, since mismatch is never negative. Nine swaps meet both conditions. After adaptation, the seven that raise the ceiling all beat the source pair, and the two that lower it gain nothing. Three of the seven switch to H5, which only adds instructions, so a revision that changes no action can still raise the ceiling through the reference it induces.

The frozen loss says little about this final gain. H0$\to$H5 and H4$\to$H3 lose the same small amount when frozen, yet H0$\to$H5 ties for the largest gain and H4$\to$H3 gains nothing. H0$\to$H1 has the largest frozen loss and still gains. The frozen return change, $T-M(\theta_0,\I_2)$, and the gain, $T-M(\theta_1,\I_2)$, differ only in the mismatch under the target harness, the target mismatch. Before adaptation, the target mismatch reflects preferences learned under the source harness and varies widely across swaps. Adaptation removes most of it, so the gain follows $T$, which differs across swaps mainly through the ceiling gain.

Ceiling and mismatch can therefore predict each gain before adaptation. Let $r_2$ be the mismatch of the checkpoint post-trained directly under the target harness, measured under that harness. If adaptation brings the target mismatch down to this same level, the gain is $T-r_2$. This prediction needs only the oracle ceilings and the six post-training runs of Section~\ref{sec:exp-training}. We recorded it before adaptation for every swap except H0$\to$H1. It gets the sign of every gain right and orders the eight recorded gains with Spearman $\rho=0.95$, while the frozen return change orders them with $\rho=0.17$. Figure~\ref{fig:adaptation} plots the gains against the frozen return change in its second panel and against $T-r_2$ in its third, and Appendix~\ref{app:exp-swaps} lists all nine swaps with their rank correlations.

\takeaway[Takeaways: What transfer costs]{Deployed return is what the harness offers (the ceiling) minus what the policy leaves unrealized (the mismatch). A harness change moves both terms, and training moves only the mismatch. A frozen comparison therefore penalizes a new harness for mismatch that adaptation later removes. All seven rejected swaps that raise the ceiling outperform the source pair after adaptation.}

\section{Measuring Mismatch in Deployment}
\label{sec:experiments}
The results so far use full action distributions and a solved optimum. A deployed agent exposes only sampled actions and may have no known optimum, so we estimate mismatch from action samples, first against a known optimal policy and then against a teacher model in its place.

\subsection{Estimating Mismatch from Action Samples}
\label{sec:probes}
Measuring mismatch requires the optimal policy, or a stand-in for it, and action samples that the deployed model draws through its unchanged interface at states on its own trajectories, since mismatch weights each decision by how often the model reaches it.

Pilot rollouts supply these states. Stopping each rollout at a random time $t$ drawn with probability $(1-\gamma)\gamma^t$ gives an exact draw from discounted occupancy. At each of $m$ such states, we sample $N$ actions through the deployed interface and smooth the counts into an action distribution $\widetilde p(\cdot\mid s)$. The mismatch estimate averages the KL divergence over the sampled states:
\begin{equation}
\widehat M=\frac{\tau}{(1-\gamma)m}\sum_{i=1}^{m}
\KL\!\left(\widetilde p(\cdot\mid s_i)\middle\Vert\pistar_\I(\cdot\mid s_i)\right).
\label{eq:sampled-mismatch}
\end{equation}

A second route needs the probabilities of only the queried actions. The scaled log odds between two actions, $\tau\log[\Pi_{\theta,\I}(a\mid s)/\Pi_{\theta,\I}(b\mid s)]$, equal the difference of their scores in the model's readout. If the queried pairs link every action to every other through a chain of pairs, these differences determine all centered scores. Comparing them with the optimal scores gives the centered score error, which bounds the mismatch.
\begin{proposition}[From score error to mismatch]
\label{thm:probe-gap}
Let $x_s=\hF_{\theta,\I}(s,\cdot)-\bF^*_\I(s,\cdot)$ and $\varepsilon_d^2=\E_{d^{\Pi_{\theta,\I}}_\I}\norm{x_s}_2^2$. For positive semantic action probabilities,
\begin{equation}
M(\theta,\I)\le\frac{\varepsilon_d^2}{4\tau(1-\gamma)}.
\label{eq:probe-gap}
\end{equation}
\end{proposition}
Once this bound falls below the threshold in Corollary~\ref{cor:adaptation}, the adapted pair is guaranteed to outperform the source pair. Appendix~\ref{app:measurement} gives the proof and a lower bound on mismatch. Section~\ref{sec:exp-prediction} tests the sampled estimate $\widehat M$ against a known optimum, and Section~\ref{sec:exp-real} applies the score-error bound with a teacher model in the optimum's place.

\subsection{Finite Measurements Predict Return Changes}
\label{sec:exp-prediction}
Section~\ref{sec:return} computed every mismatch exactly. We now ask whether the sampled estimate $\widehat M$ of Eq.~\eqref{eq:sampled-mismatch} can replace it, both for large return changes and for the small differences that decide whether adaptation pays off. An oracle supplies the ceiling and the optimal action distributions, and the deployed model is accessed only through sampled actions. The predicted return change of a candidate relative to its baseline is $\widehat{\Delta J}=\Delta C-\Delta\widehat M$, and exact returns on tasks and configurations held out from calibrating the estimator serve as ground truth. We compare this prediction with two baselines at the same query budget. The ceiling alone sets $\Delta\widehat M=0$. A frequency rollout estimates each return directly by running the policy and estimating each selected action's log probability from repeated queries.

\begin{figure}[t]
\begin{minipage}[c]{0.66\linewidth}
\includegraphics[width=\linewidth]{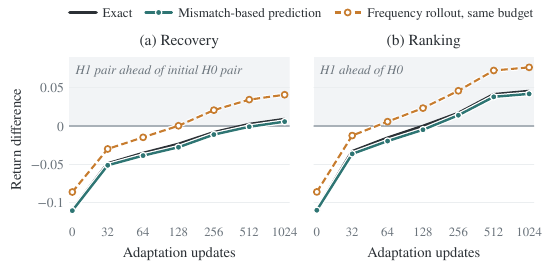}
\end{minipage}\hfill
\begin{minipage}[c]{0.31\linewidth}
\caption{Sampled mismatch tracks the small return differences that decide adaptation. Curves show exact values and the two predictions at each saved H1-adapted checkpoint. (a) Recovery, the checkpoint's return under H1 minus the source H0 pair's return. (b) Harness ranking, the checkpoint's return under H1 minus its return under H0.}
\label{fig:diagnostic-adaptation}
\end{minipage}
\vspace{-10px}
\end{figure}

\paragraph{Large changes across configurations.}
We deploy the final H0-trained and H1-adapted checkpoints through H0's native actions and vary one aspect of the deployment at a time, namely how reliably a click executes, what a click costs, or how the prompt is presented. We predict each configuration's return change relative to a fixed baseline configuration. The ceiling alone fails, because presentation changes move the mismatch and leave the ceiling fixed. Subtracting the sampled mismatch reduces the prediction error about a hundredfold and gets the direction of every change not close to zero. The frequency rollout is about as accurate, so on large changes the two estimators perform alike.

\paragraph{Small differences during adaptation.}
Recovery and harness ranking from Section~\ref{sec:adaptation} are differences of a few hundredths in return, and their signs decide whether adaptation pays off. We predict both at each saved H1-adapted checkpoint. Here the mismatch-based prediction is about ten times more accurate than the frequency rollout. It takes the ceiling difference from the oracle and estimates only the mismatch, whereas the rollout estimates the full return of each pair. The predicted ranking reverses at the same checkpoint as the observed one, and the predicted recovery comes one saved checkpoint later. The frequency rollout overestimates both differences and predicts the reversal too early. Figure~\ref{fig:diagnostic-adaptation} shows both predictions, and Appendices~\ref{app:exp-measurement} and~\ref{app:exp-prediction} give the sampling budget and all numerical results.

\subsection{Ceiling and Mismatch of Coding-Agent Harnesses}
\label{sec:exp-real}
We deploy a frozen Qwen3.5-9B through the native configurations of five coding-agent harnesses: DeepAgents, Pi coding-agent, DeepSeek Harness, Terminus-2, and mini-SWE-agent. With the model fixed, the harness moves Terminal-Bench Lite accuracy from 47.9\% to 64.1\%. We ask whether the ceiling and the mismatch explain this ordering.

\begin{wrapfigure}[31]{l}{0.5\linewidth}
\vspace{-12pt}
\begin{minipage}{\linewidth}
\centering\small
\makeatletter\def\@captype{table}\makeatother
\setlength{\abovecaptionskip}{0pt}
\setlength{\belowcaptionskip}{4pt}
\caption{Ceiling and mismatch of five coding-agent harnesses for a frozen Qwen3.5-9B. $\widetilde C$ and $\widetilde M$ are teacher-based estimates. The last row gives each column's Spearman correlation with accuracy.}
\label{tab:real-agents}
\setlength{\tabcolsep}{2.5pt}
\begin{tabular}{@{}lrrrr@{}}
\toprule
Harness & $\widetilde C$ & $\widetilde M$ & $\widetilde C-\widetilde M$ & \shortstack{Accuracy\\(\%)}\\
\midrule
Terminus-2 & 0.431 & 0.134 & 0.297 & $64.1\pm5.1$\\
mini-SWE-agent & 0.631 & 0.335 & 0.296 & $59.8\pm7.8$\\
DeepSeek Harness & 0.590 & 0.382 & 0.208 & $54.7\pm3.9$\\
Pi coding-agent & 0.383 & 0.206 & 0.177 & $51.3\pm2.6$\\
DeepAgents & 0.261 & 0.429 & $-0.168$ & $47.9\pm3.9$\\
\midrule
Spearman $\rho$ & 0.7 & $-0.7$ & 1.0 & \\
\bottomrule
\end{tabular}
\end{minipage}

\vspace{6pt}
\centering
\includegraphics[width=0.92\linewidth,trim=212 16 0 2,clip]{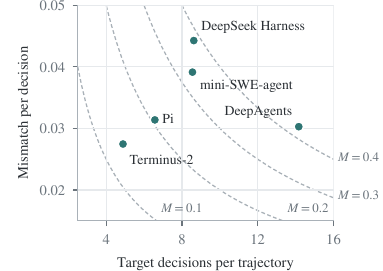}
\caption{Mismatch accumulates over decisions. Each harness's mismatch splits into the number of decisions the evaluated model makes per trajectory and the mismatch per decision. Dashed curves have constant $\widetilde M$.}
\label{fig:real-agents}
\end{wrapfigure}
\paragraph{Teacher-based estimates.}
Coding agents have open action spaces and no solvable optimum, so a stronger teacher model, GPT-5.6-luna, takes the optimum's place. Below, the evaluated model means the frozen Qwen3.5-9B. On three Terminal-Bench Lite tasks, we record every decision of the evaluated model, sample outputs from it and from the teacher through the unchanged harness, and map each output to an action category shared by all harnesses, such as inspecting files, modifying them, running commands, or planning. We use the evaluated model's own category distribution as the reference. Its KL penalty is then zero, so its regularized return equals its success probability, and by Eq.~\eqref{eq:ceiling-mismatch} so does its exact ceiling minus mismatch. We estimate the ceiling by the teacher's regularized return, $\widetilde C$. We estimate the mismatch, $\widetilde M$, with the bound of Proposition~\ref{thm:probe-gap}, using the teacher's log odds in place of the optimal scores and summing along the evaluated model's trajectories. Because the teacher is not optimal, we test whether the estimates order the harnesses by accuracy. Appendix~\ref{app:exp-realagents} gives the protocol and estimators.

\paragraph{$\widetilde C-\widetilde M$ orders the harnesses.}
Computed without the evaluated model's task outcomes, $\widetilde C-\widetilde M$ orders all five harnesses as accuracy does. Either term alone misorders them. The ceiling estimate alone ranks mini-SWE-agent first, and the mismatch estimate alone ranks Pi coding-agent second, while their difference ranks both correctly. \mbox{Terminus-2} has a middle ceiling and leads through the smallest mismatch. DeepSeek Harness and mini-SWE-agent have the two highest ceilings and large mismatch, the profile of the batched-click harness before adaptation. Table~\ref{tab:real-agents} lists the estimates.

The same pattern appears when both terms are exact. On MiniWoB++, where they can be computed, we let a frozen \mbox{Qwen3.5-9B} serve as its own reference under H0--H5. The ceiling alone then ranks the most successful harness last, and the mismatch alone ranks H0 below two less successful harnesses, while only their difference orders all six correctly. Appendix~\ref{app:exp-selection} gives this comparison.

\paragraph{Where mismatch comes from.}
By Proposition~\ref{thm:gap}, mismatch accumulates along the model's own trajectories, so it grows with the number of decisions a harness asks of the model. DeepAgents has a mismatch per decision close to that of \mbox{Terminus-2}, but the evaluated model makes nearly three times as many model calls under it and accumulates the largest mismatch. DeepSeek Harness has the largest mismatch per decision. Under it, the teacher plans much more often than the evaluated model, and the evaluated model inspects the environment much more often. Figure~\ref{fig:real-agents} separates the two factors for each harness.

\section{Discussion and Conclusion}
\label{sec:conclusion}
A policy trained under a harness learns the harness's relative action scores and expresses them through its interface. Post-training makes these preferences harness-specific, and a new interface can change the model's choices even when every optimal decision stays the same. Because deployed return is the ceiling minus the mismatch, a harness that offers more can hurt a trained model and become the best choice after adaptation. Mismatch estimated from sampled actions predicts these changes when the optimum is known, and for coding agents, teacher-based estimates of ceiling minus mismatch order harnesses by accuracy. Harness evaluations should therefore report the ceiling next to the frozen return, evaluate several checkpoints under each harness rather than a single one, and check how presentation changes move the model's choices. Harness distillation and joint adaptation of harnesses and weights \citep{ye2026harnesszero,chen2026coharness,kim2026whale} can use sampled-action measurements to track the ceiling and the mismatch separately. Appendix~\ref{app:related} discusses related work.

\label{main-text-end}
\begingroup
\let\emph\textup
\bibliography{workspace/iclr2027_conference}
\endgroup
\bibliographystyle{iclr2027_conference}
\clearpage

\setlength{\textfloatsep}{20pt plus 2pt minus 4pt}
\setlength{\abovecaptionskip}{10pt}
\appendix
\section{Related Work}
\label{app:related}
\label{sec:related}
\textbf{Regularized control and decision geometry.} Control as inference and RL with a maximum entropy objective express optimal policies through rewards, action values, and reference distributions \citep{levine2018inference,haarnoja2017reinforcementlearningdeepenergybased,haarnoja2018softactorcriticoffpolicymaximum}. Regularized control supplies the Gibbs policy and the geometry of the value gap \citep{geist2019regularized,odonoghue2022bregman}. The relation between policy gradients and soft Q-learning provides another view of these coordinates \citep{schulman2018equivalencepolicygradientssoft}. DPO and its sequential extension relate language-model policies to rewards and action values during post-training \citep{rafailov2023dpo,rafailov2024q}.

Our optimal action score combines the optimal semantic action value with the semantic reference prior. These coordinates let us compare the decisions induced by different interfaces and measure a deployed policy's gap to each optimum. The analysis of fine-tuning objectives by \citet{swamy2025allroads} likewise uses a common geometric description to distinguish ideal equivalence from differences that emerge in practice. Here the distinction concerns the interface's optimal decisions and the choices that a particular checkpoint makes through it.

\textbf{Behavioral identification and reward transfer.} Inverse RL recovers rewards from behavior subject to ambiguities \citep{Ng2000Algorithms,russell1998colt-learning}. Reward shaping using a potential function preserves optimal policies \citep{Ng1999Policy}. Analyses of identifiability under entropy regularization and of partial identifiability characterize the corresponding reward classes \citep{cao2021identifiability,skalse2023invariancepolicyoptimisationpartial}. EPIC compares reward functions while accounting for transformations that preserve policies \citep{gleave2021quantifyingdifferencesrewardfunctions}. Work on the transferability of rewards recovered through regularized inverse RL further studies changes in transition laws \citep{schlaginhaufen2024transferability}.

Our equivalence class for a single interface removes statewise constants from the scores. Appendix~\ref{app:identification} distinguishes this behavioral object from the nonunique decomposition into prior, reward, and transition. The transfer analysis then compares the target energy with the actual deployed readout, weighting their discrepancy by deployment occupancy.

\textbf{Interface and agent design.} Interface design and prompt formatting affect agent behavior \citep{yang2024sweagent,sclar2024format,zhang2025ufo}. GUI-360$^\circ$ reports higher action prediction accuracy for GPT-4o and lower accuracy for fine-tuned Qwen after adding accessibility metadata and replacing coordinate arguments with element IDs \citep{mu2025gui360}. ADAS and AFlow search over agent implementations and workflows \citep{hu2025adas,zhang2025aflow}. Darwin G\"odel Machine and Self-Improving Coding Agent revise scaffolds and assess them through execution \citep{zhang2026darwingodelmachineopenended,robeyns2025selfimprovingcodingagent}. Promptbreeder, ToolMaker, and MAS-ZERO respectively explore prompts, tools, and designs involving multiple agents \citep{fernando2023promptbreederselfreferentialselfimprovementprompt,wolflein-etal-2025-llm,ke2025maszero}. Surveys place these approaches within agent systems with mechanisms for self improvement \citep{fang2025comprehensivesurveyselfevolvingai,gao2026surveyselfevolvingagentswhat}.

These design procedures can change action opportunities, realizations, or both. Our analysis separates each candidate's optimal return from the evaluated checkpoint's gap to that optimum. This distinction connects execution feedback to two possible sources of improvement: a more useful decision problem and a policy that better realizes its available decisions.

\textbf{Context and harness distillation.} Context distillation trains a model to reproduce, without the prompt, the behavior that the prompt elicits \citep{snell2022context}. Harness-Zero distills a specialized harness into the weights by letting a harnessing agent correct the student's responses in the target harness's action space \citep{ye2026harnesszero}. Our identification result states which decision information such training can carry, the relative action scores of the training harness. The readout study shows that the learned preferences are still expressed through the deployment interface, so a distilled policy can change its choices when the interface changes.

\textbf{Joint adaptation of harnesses and model weights.} Joint adaptation takes several forms: SIA selects harness and weight updates, Co-Harness trains from trajectories collected under revised interfaces, and WHALE alternates training with harness search \citep{hebbar2026sia,chen2026coharness,kim2026whale}. \citet{yu2026coevolving} find that imitating expert trajectories can hurt a weaker model under an evolved harness while on-policy correction helps, and HarnessBandit schedules training across harnesses by learnability and transferability \citep{wei2026harnessbandit}.

These methods can use the ceiling and the mismatch to tell whether a candidate harness offers too little or is not yet used well. Our adaptation threshold specifies the mismatch reduction needed to make a revision with a higher ceiling beneficial. Eq.~\eqref{eq:adaptation-interaction} isolates how a weight update changes a revision's relative value. Comparisons across checkpoints and interfaces therefore track both recovery after a revision and changes in interface rankings. The controlled protocol measures this interaction separately from gains under either interface alone.

\section{Semantic Control: Objects, Abstraction, and Objectives}
\label{app:semantics}
\subsection{Notation and Standing Conditions}
\begin{table}[h]
\centering\small
\caption{Notation. Energies and vector operations are statewise unless an expectation is shown.}
\begin{tabularx}{\linewidth}{lY}
\toprule
Symbol & Meaning\\
\midrule
$\I=(c_\I,d_\I,e_\I)$ & Context encoder, semantic decoder, and fixed execution protocol.\\
$\pi_\theta,\ \Pi_{\theta,\I}$ & Policy over token sequences and its executed semantic pushforward.\\
$P_\I,r_\I,\piref_\I$ & Induced dynamics, task reward including execution costs, and semantic reference prior.\\
$\rho_0,\gamma,\tau$ & Initial state distribution, discount, and KL regularization coefficient.\\
$F^*_\I,\ \bF^*_\I$ & Optimal action score and its centered representative.\\
$\hF_{\theta,\I}$ & Readout, the centered action scores $\tau\proj\log\Pi_{\theta,\I}$.\\
$d^\Pi_\I$ & Normalized discounted state occupancy of the evaluated policy.\\
$C(\I),M(\theta,\I)$ & Ceiling (optimal regularized return) and mismatch.\\
$e_1,g_{12},\eta_{12}$ & Training residual, target shift, and readout drift.\\
$\Phi,D_\Phi$ & Log partition function and its Bregman divergence.\\
\bottomrule
\end{tabularx}
\end{table}
The formal results use finite state and action spaces, bounded rewards, $0\le\gamma<1$, $\tau>0$, and a reference prior that is positive on every available action. The initial distribution, reward units, and regularization coefficient are shared in a comparison across interfaces. State/action alignments are fixed before the comparison. Results involving $\log\Pi_{\theta,\I}$ additionally require positive probabilities for the actual policy on the probed actions. The gap identity for a fixed interface itself allows zero policy probabilities, with $0\log0=0$.

We use a reference checkpoint distinct from the evolving evaluated checkpoint. When its sequence distribution is $\pi_{\rm ref}$, a natural semantic reference is
\begin{equation}
\piref_\I(a\mid s)=\sum_{y:d_\I(y;s)=a}\pi_{\rm ref}(y\mid c_\I(s)).
\label{eq:reference-pushforward}
\end{equation}
Consequently a new encoding can change both the target induced by the reference and the evaluated model's readout. Experiments that hold the target fixed must verify or explicitly control this reference.

\subsection{Decision Epochs and Controller State}
A semantic state includes every variable needed to predict the state at the next decision: the task state, visible history or a sufficient summary, tool results, budget counters, and controller memory. An action denotes a fully specified operation, such as invoking a particular tool with particular arguments, or submitting a particular candidate. Two output strings belong to one semantic action only when the decoder and executor execute the same operation at that state.

After a semantic action is selected, $e_\I$ can execute a fixed tool, perform a deterministic check, or apply a stochastic transition independent of the policy. Its expected costs enter $r_\I$, and its outcomes enter $P_\I$. If a retry, verifier, or stopping rule calls the evaluated model for another decision, that call is another decision epoch in the expanded state space. This keeps the transition kernel fixed while the policy varies.

An invalid string is mapped to an explicit failure or recovery action. Similarly, when an empirical action vocabulary groups outputs into an ``other'' category, that category must have a defined execution interpretation. Discarding unmodeled outputs changes the policy being measured.

\subsection{When Training over Tokens Induces Semantic Optimality}
Let $Y_a(s)$ be the set of complete output realizations decoded as action $a$. Write a sequence policy and reference as
\[
p(y\mid s)=p_A(a\mid s)p(y\mid s,a),\qquad
\rho(y\mid s)=\rho_A(a\mid s)\rho(y\mid s,a),\quad y\in Y_a(s).
\]
For any such distributions, the KL chain rule gives
\begin{equation}
\KL(p(\cdot\mid s)\|\rho(\cdot\mid s))=
\KL(p_A(\cdot\mid s)\|\rho_A(\cdot\mid s))+
\E_{a\sim p_A}\KL(p(\cdot\mid s,a)\|\rho(\cdot\mid s,a)).
\label{eq:kl-chain}
\end{equation}
Suppose reward, transition, and discount are identical within each fiber. Then the return from choosing $y$ depends only on $a$, while the second term in Eq.~\eqref{eq:kl-chain} is minimized by retaining the reference conditional distribution inside that fiber. Optimization over complete sequence actions therefore reduces exactly to the semantic control problem regularized by KL divergence relative to $\rho_A$.

For an autoregressive model, probabilities of complete sequences include termination. The expected sum of token KL divergences equals sequence KL when the same terminating sequence measure and matching token conditionals are used. Rewards that depend on tokens, discounts that depend on sequence length, or additional model decisions within a proposed temporally extended action require the expanded model above.

\section{Behavioral Identification and Mechanism Ambiguity}
\label{app:identification}
\subsection{Soft Bellman Optimality}
For a fixed interface, the optimal Bellman operator is
\[
(\mathcal TV)(s)=\tau\log\sum_a\piref(a\mid s)
\exp\left(\frac{r(s,a)+\gamma\E_{P(\cdot\mid s,a)}V(s')}{\tau}\right).
\]
Monotonicity of the log partition function and its translation identity imply
$\norm{\mathcal TV-\mathcal TW}_\infty\le\gamma\norm{V-W}_\infty$.
Thus it has a unique fixed point $V^*$, and $Q^*=r+\gamma PV^*$ is uniquely determined. At each state, maximizing
$\sum_a\pi_a Q_a^*-\tau\sum_a\pi_a\log(\pi_a/\rho_a)$
over the simplex gives
\[
\pi_a^*=\frac{\rho_a e^{Q_a^*/\tau}}{\sum_b\rho_b e^{Q_b^*/\tau}}.
\]
Strict concavity gives uniqueness. At the fixed point, $V^*(s)=\tau\log\sum_b\rho_b e^{Q_b^*/\tau}$, so the optimal policy takes the Gibbs form
\begin{equation}
\pistar(a\mid s)=\piref(a\mid s)\exp\!\left(\frac{Q^*(s,a)-V^*(s)}{\tau}\right).
\label{eq:gibbs-policy}
\end{equation}
Substituting $F_a^*=Q_a^*+\tau\log\rho_a$ produces the softmax representation in the main text.

\subsection{Proof of Proposition~\ref{thm:identification}}
If $F(s,\cdot)=G(s,\cdot)+c(s)\one$, the factor $e^{c(s)/\tau}$ cancels in softmax. Conversely, equal softmax probabilities imply
\[
F(s,a)-\Phi(F(s,\cdot))=G(s,a)-\Phi(G(s,\cdot))
\]
for every action. Their difference is therefore constant across actions. Applying this equivalence to optimal action scores proves the first assertion. Under $\Pi_{\theta,\I_{\rm tr}}=\pistar_{\I_{\rm tr}}$, the Gibbs identity gives
\[
\tau\log\Pi_{\theta,\I_{\rm tr}}(a\mid s)=F^*_{\I_{\rm tr}}(s,a)-V^*_{\I_{\rm tr}}(s).
\]
Projection onto $\one^\perp$, or subtraction between two actions, removes $V^*_{\I_{\rm tr}}(s)$ and proves Eq.~\eqref{eq:identification}. \qed

\subsection{Mechanism Nonidentifiability}
\label{app:mechanism}
\begin{proposition}[Mechanism nonidentifiability]
\label{prop:nonidentification}
A fixed optimal score field can be realized by a continuous family of different reference priors and rewards, and even by different transition laws. Consequently, optimal behavior alone does not uniquely determine the interface mechanism.
\end{proposition}
\paragraph{Proof.}
Fix a finite score field $F$, any reference $\rho'$ with full support, and any stochastic kernel $P'$. Define
\[
V(s)=\Phi(F(s,\cdot)),\qquad Q'(s,a)=F(s,a)-\tau\log\rho'(a\mid s),
\qquad r'=Q'-\gamma P'V.
\]
Then $Q'=r'+\gamma P'V$, and
\[
\tau\log\sum_a\rho'(a\mid s)e^{Q'(s,a)/\tau}
=\tau\log\sum_a e^{F(s,a)/\tau}=V(s).
\]
By uniqueness of the Bellman fixed point, these are the optimal values of the constructed control problem. Its optimal policy is $\sm(F/\tau)$. Varying $\rho'$ or $P'$ gives a continuous family of mechanisms with the same scores and policy. \qed

\section{Performance Geometry and Readout Across Interfaces}
\label{app:geometry}
\subsection{Proof of Proposition~\ref{thm:gap}}
Fix an interface and suppress its subscript. For a policy $\Pi$, write
\[
(\mathcal T^\Pi V)(s)=\E_{a\sim\Pi(\cdot\mid s)}\left[r(s,a)+\gamma PV(s,a)-\tau\log\frac{\Pi(a\mid s)}{\piref(a\mid s)}\right].
\]
Rearranging the optimal-policy expression in Eq.~\eqref{eq:gibbs-policy} gives $Q^*-V^*=\tau\log(\pistar/\piref)$. Substituting this relation into the regularized Bellman update yields the statewise gap:
\begin{equation}
V^*(s)-(\mathcal T^\Pi V^*)(s)=\tau\KL(\Pi(\cdot\mid s)\|\pistar(\cdot\mid s))=:k_\Pi(s).
\label{eq:statewise-gap}
\end{equation}
Subtracting the policy Bellman equation $V^\Pi=\mathcal T^\Pi V^\Pi$ yields
\[
V^*-V^\Pi=k_\Pi+\gamma P^\Pi(V^*-V^\Pi)
=(I-\gamma P^\Pi)^{-1}k_\Pi.
\]
The normalized occupancy satisfies
$(d^\Pi)^\top=(1-\gamma)\rho_0^\top(I-\gamma P^\Pi)^{-1}$.
Multiplication by $\rho_0^\top$ proves Eq.~\eqref{eq:gap}, in which both the KL direction and the occupancy belong to the evaluated policy. \qed

\subsection{Bregman Representation of the Decision Loss}
\label{app:bregman}
For $\Phi(F)=\tau\log\sum_a e^{F_a/\tau}$, the following identity relates score differences to the policy divergence.
\begin{lemma}[Bregman--KL identity]
\label{lem:bregman-kl}
For any score vectors $F$ and $G$, the Bregman divergence generated by $\Phi$ satisfies
\begin{equation}
D_\Phi(F,G)=\Phi(F)-\Phi(G)-\inner{\nabla\Phi(G)}{F-G}
=\tau\KL\bigl(\sm(G/\tau)\|\sm(F/\tau)\bigr).
\label{eq:bregman}
\end{equation}
\end{lemma}
\paragraph{Proof.}
Let $p_F=\sm(F/\tau)$, so that $\nabla\Phi(F)=p_F$ and $\tau\log p_F=F-\Phi(F)\one$. Then
\begin{align*}
\tau\KL(p_G\|p_F)
&=\sum_a(p_G)_a\bigl[G_a-\Phi(G)-F_a+\Phi(F)\bigr]\\
&=\Phi(F)-\Phi(G)-\inner{p_G}{F-G}=D_\Phi(F,G).
\end{align*}
For any constants $c,d$, $\Phi(F+c\one)=\Phi(F)+c$ and $\nabla\Phi(G+d\one)=\nabla\Phi(G)$. These identities show $D_\Phi(F+c\one,G+d\one)=D_\Phi(F,G)$. For a positive policy, let $F_\Pi=\tau\proj\log\Pi$. Substitution in Proposition~\ref{thm:gap} gives the performance gap in centered scores:
\begin{equation}
J_\I(\pistar_\I)-J_\I(\Pi)
=\frac{1}{1-\gamma}\E_{s\sim d^\Pi_\I}D_\Phi(F^*_\I,F_\Pi).
\label{eq:energy-gap}
\end{equation}
For $\Pi=\Pi_{\theta,\I}$, its centered score vector is $\hF_{\theta,\I}$, giving the score form of mismatch through Eq.~\eqref{eq:local-energy-loss}.

\subsection{General Transfer and the Case with Zero Drift}
Eq.~\eqref{eq:three-errors} follows by adding and subtracting the source target and source readout in the same centered coordinates. Substitution into the exact gap gives
\begin{equation}
M(\theta,2)=\frac{1}{1-\gamma}\E_{d^{\Pi_{\theta,2}}_2}
D_\Phi\bigl(\bF^*_2,\bF^*_1+e_1+\eta_{12}\bigr).
\label{eq:general-transfer}
\end{equation}
For an optimal source policy whose readout is preserved by the swap, $e_1=\eta_{12}=0$, so
\begin{equation}
M(\theta,2)=\frac{1}{1-\gamma}\E_{d^{\Pi_{\theta,2}}_2}D_\Phi(F^*_2,F^*_1)
\qquad(e_1=0,\ \eta_{12}=0).
\label{eq:zero-drift}
\end{equation}
The model continues to express the source preferences, whose discrepancy from the new optimum determines the deployment gap. For a quantitative perturbation bound, define $\varepsilon_g,\varepsilon_e,\varepsilon_\eta$ as the respective root mean square norms of $g_{12},e_1,\eta_{12}$ under the same occupancy of the deployed policy in the target interface. Proposition~\ref{thm:probe-gap} and Minkowski's inequality give
\begin{equation}
M(\theta,2)\le\frac{(\varepsilon_g+\varepsilon_e+\varepsilon_\eta)^2}{4\tau(1-\gamma)}.
\label{eq:transfer-bound}
\end{equation}

\subsection{Correspondence Controls Readout Drift}
\label{app:correspondence}
To relate readout stability to model behavior, suppose $T_s$ bijectively matches output realizations of each semantic action across interfaces. Define the model's sensitivity to this correspondence by
\begin{equation}
\delta_{\rm tok}(s)=\sup_y\left|\log\pi_\theta(T_s y\mid c_2(s))-\log\pi_\theta(y\mid c_1(s))\right|.
\label{eq:token-distortion}
\end{equation}
\begin{proposition}[Correspondence controls readout drift]
\label{thm:drift}
For a bijection that preserves semantic action fibers, with finite $\delta_{\rm tok}(s)$,
\begin{equation}
|\eta_{12}(s,a)-\eta_{12}(s,b)|\le 2\tau\delta_{\rm tok}(s),
\qquad \norm{\eta_{12}(s,\cdot)}_2\le\tau\sqrt{n}\,\delta_{\rm tok}(s).
\label{eq:drift-bound}
\end{equation}
\end{proposition}
\paragraph{Proof.}
At a fixed state, let $p_i(y)=\pi_\theta(y\mid c_i(s))$ and $\delta=\delta_{\rm tok}(s)$. The bijection that preserves action fibers gives
\[
\Pi_{\theta,2}(a\mid s)=\sum_{y\in Y_a^{(1)}(s)}p_2(T_s y).
\]
Since $e^{-\delta}p_1(y)\le p_2(T_sy)\le e^\delta p_1(y)$, summing over the fiber yields
\[
e^{-\delta}\Pi_{\theta,1}(a\mid s)\le\Pi_{\theta,2}(a\mid s)
\le e^\delta\Pi_{\theta,1}(a\mid s).
\]
Write $v_a=\log\Pi_{\theta,2}(a\mid s)-\log\Pi_{\theta,1}(a\mid s)$. Then $|v_a|\le\delta$, and $\eta_{12}=\tau\proj v$. Pairwise subtraction gives $|\eta_a-\eta_b|\le2\tau\delta$. Orthogonal projection is nonexpansive, so
$\norm{\eta_{12}}_2\le\tau\norm v_2\le\tau\sqrt n\,\delta$. \qed

\subsection{A Two-Action Illustration of Reversal and Adaptation}
\label{app:bandit-reversal}
Consider a bandit with two actions and a uniform reference. The source execution protocol gives rewards $(1,0)$, and its optimal policy chooses the first action with probability $p=e^{1/\tau}/(1+e^{1/\tau})$. A revised protocol lowers the first action's utility to $0$ and raises the second action's utility to $u>0$. Holding the semantic policy fixed isolates the effect of this change in action values.

The two ceilings are
\[
C_1=\tau\log\frac{e^{1/\tau}+1}{2},\qquad
C_2=\tau\log\frac{1+e^{u/\tau}}{2}.
\]
Hence $C_2>C_1$ if and only if $u>1$. Holding the actual policy fixed also holds its reference KL fixed. Its returns satisfy
\[
J_2-J_1=u(1-p)-p.
\]
The policy gains $u$ when it takes the second action and loses $1$ when it takes the first. Its return therefore improves only when $u>p/(1-p)=e^{1/\tau}$. Throughout $1<u<e^{1/\tau}$, the revised interface has a higher ceiling and a lower deployed return: the unchanged policy still favors the action preferred under the source optimum.

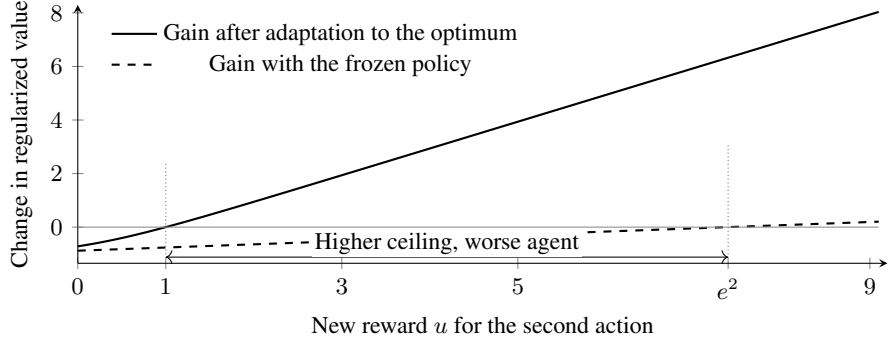
\begin{figure}[htbp]
\centering\begin{tikzpicture}
\begin{axis}[
width=.88\linewidth,height=5.0cm,
xmin=0,xmax=9.2,ymin=-1.35,ymax=8.3,
xlabel={New reward $u$ for the second action},ylabel={Change in regularized value},
axis lines=left,tick label style={font=\small},label style={font=\small},
xtick={0,1,3,5,7.389,9},xticklabels={$0$,$1$,$3$,$5$,$e^2$,$9$},
ytick={0,2,4,6,8},
legend style={draw=none,at={(.03,.97)},anchor=north west,font=\small},
clip=true]
\addplot[black,thick,domain=0:9.1,samples=100]{.5*ln((1+exp(2*x))/(1+exp(2)))};
\addlegendentry{Gain after adaptation to the optimum}
\addplot[black,thick,dashed,domain=0:9.1,samples=100]{(x-exp(2))/(1+exp(2))};
\addlegendentry{Gain with the frozen policy}
\draw[black!45] (axis cs:0,0)--(axis cs:9.1,0);
\draw[densely dotted,black!55] (axis cs:1,-1.15)--(axis cs:1,2.4);
\draw[densely dotted,black!55] (axis cs:7.389,-1.15)--(axis cs:7.389,3.1);
\draw[<->] (axis cs:1,-1.12)--node[above,font=\small,fill=white,inner sep=1pt]{Higher ceiling, worse agent}(axis cs:7.389,-1.12);
\end{axis}
\end{tikzpicture}
\caption{Adaptation can realize an initially harmful interface improvement. Analytical curves for the two-action bandit with $\tau=0.5$. The frozen policy exceeds the original pair only for $u>e^2$ (dashed), while the ceiling gain is positive for $u>1$. Reaching the new optimum realizes the ceiling gain $C_2-C_1$ (solid). Corollary~\ref{cor:adaptation} gives the mismatch threshold for a benefit before full adaptation.}
\label{fig:analytic-reversal}
\end{figure}

Because the source policy is optimal, Corollary~\ref{cor:adaptation} makes the revision beneficial once the remaining mismatch is smaller than $C_2-C_1$. Reaching the new optimum realizes this entire gain, positive for every $u>1$. Figure~\ref{fig:analytic-reversal} plots the frozen policy's return change and the ceiling gain at $\tau=0.5$, with the vertical distance between the curves giving the mismatch under the revised interface. All quantities here are analytical.

\subsection{Policy Adaptation and Interface Rankings}
\label{app:adaptation}
\paragraph{Proof of Corollary~\ref{cor:adaptation}.}
Keep the evaluation objective and reference fixed for each interface. Substituting $J(\theta,\I_j)=C_j-M(\theta,\I_j)$ gives
\begin{align*}
J(\theta_0,\I_2)-J(\theta_0,\I_1)
&=\Delta C+M(\theta_0,\I_1)-M(\theta_0,\I_2),\\
J(\theta_1,\I_2)-J(\theta_0,\I_1)
&=\Delta C+M(\theta_0,\I_1)-M(\theta_1,\I_2).
\end{align*}
The initially harmful revision makes the first difference negative. Since $\Delta C+M(\theta_0,\I_1)=C_2-J(\theta_0,\I_1)$, the second is positive exactly under Eq.~\eqref{eq:adaptation-threshold}. Combining the initial harm and subsequent improvement gives the equivalent chain
\[
M(\theta_1,\I_2)<\Delta C+M(\theta_0,\I_1)<M(\theta_0,\I_2).
\]
The source pair's mismatch $M(\theta_0,\I_1)$ allows for a source checkpoint that has not reached its optimum. For an optimal source policy, the threshold equals the ceiling gain $\Delta C$, and reaching the new optimum realizes the entire gain. \qed

\paragraph{Interaction between weight and interface updates.}
For $j\in\{1,2\}$, define the gain from the weight update in interface $j$ by
\[
A_j=J(\theta_1,\I_j)-J(\theta_0,\I_j)
=M(\theta_0,\I_j)-M(\theta_1,\I_j).
\]
Write $\Delta J(\theta)=J(\theta,\I_2)-J(\theta,\I_1)$ for the relative value of the revision. The interaction across the four combinations is
\begin{equation}
\Omega=\Delta J(\theta_1)-\Delta J(\theta_0)=A_2-A_1.
\label{eq:adaptation-interaction}
\end{equation}
The ceiling terms cancel because the objectives are fixed across checkpoints. A positive $\Omega$ means the revision becomes relatively more valuable after the weight update. If $\Delta J(\theta_0)<0$, its ranking reverses exactly when $\Omega>-\Delta J(\theta_0)$. This compares the two interfaces at the updated checkpoint, while Corollary~\ref{cor:adaptation} compares the adapted pair with the source pair. Reporting $A_1$ and $A_2$ also identifies the source of the interaction: deterioration under the source interface ($A_1<0$) can increase $\Omega$ even when $A_2=0$.

\section{Measurement Theory}
\label{app:measurement}
\subsection{Curvature on the Decision Quotient}
For $p=\sm(F/\tau)$,
\begin{equation}
\nabla^2\Phi(F)=\frac{1}{\tau}\bigl(\diag(p)-pp^\top\bigr).
\label{eq:hessian}
\end{equation}
The categorical covariance annihilates $\one$. Its spectral norm is at most $1/2$: the absolute row sum of row $i$ is $2p_i(1-p_i)\le1/2$, and the matrix is symmetric. If $p_i\ge p_{\min}$ and $z\perp\one$, then
\begin{align*}
z^\top\bigl(\diag(p)-pp^\top\bigr)z
&=\min_c\sum_i p_i(z_i-c)^2\\
&\ge p_{\min}\min_c\sum_i(z_i-c)^2
=p_{\min}\norm z_2^2.
\end{align*}
Thus the upper Hessian constant is $1/(2\tau)$ and the lower constant on the centered subspace is $p_{\min}/\tau$.

\subsection{Bounds on the Performance Gap}
\label{app:probe-bounds}
Proposition~\ref{thm:probe-gap} bounds mismatch through centered score error. An upper bound below the threshold in Corollary~\ref{cor:adaptation} guarantees recovery of the source pair's return. Under a condition on probability overlap, the same score error also provides a lower bound on mismatch. We state the complementary bound and prove both results below.
\begin{corollary}[Complementary lower bound]
\label{cor:probe-lower}
With $x_s$ and $\varepsilon_d$ as in Proposition~\ref{thm:probe-gap}, let $p_{\min}>0$ be a uniform lower bound on action probabilities along the score interpolation between the actual and optimal policies. Then
\begin{equation}
\frac{p_{\min}\varepsilon_d^2}{2\tau(1-\gamma)}\le M(\theta,\I)
\le\frac{\varepsilon_d^2}{4\tau(1-\gamma)}.
\label{eq:probe-gap-two-sided}
\end{equation}
\end{corollary}
\paragraph{Proof of both bounds.}
At a fixed state, let $x=\hF-\bF^*$, with both arguments centered. Taylor's integral remainder gives
\[
D_\Phi(\bF^*,\hF)=\int_0^1(1-t)x^\top\nabla^2\Phi\bigl(\hF+t(\bF^*-\hF)\bigr)x\,\mathrm dt.
\]
The curvature bounds above and $\int_0^1(1-t)\,\mathrm dt=1/2$ imply
\[
\frac{p_{\min}}{2\tau}\norm x_2^2
\le D_\Phi(\bF^*,\hF)
\le\frac{1}{4\tau}\norm x_2^2.
\]
Taking the actual occupancy expectation and dividing by $1-\gamma$ proves both bounds. \qed

A lower probability bound at both endpoint policies is sufficient along their score interpolation: the logarithm of softmax is concave in the score vector, so each interpolated log probability is at least the linear interpolation of its endpoint log probabilities.

The quadratic upper rate and constant are locally tight. For two actions around the uniform distribution, the categorical covariance has eigenvalue $1/2$ on $\one^\perp$, so $D_\Phi(F,F+z)=\norm z_2^2/(4\tau)+o(\norm z_2^2)$ for centered $z$. No positive quadratic lower constant independent of the action probabilities is possible: near a policy concentrated on one action, the nonconstant curvature can approach zero.

\section{Experimental Details}
\label{app:experiments}
We describe the experiments in the order of the main text: the MiniWoB++ setup and post-training study, the readout study, the H0-to-H1 adaptation study, the adaptation across harness swaps, the calibration of finite measurements, the independent prediction study, the coding-agent harness study, and the exact comparison of harness scores with task success.

\subsection{MiniWoB++ Setup and Post-Training}
\label{app:e1}
The controlled study pairs language-model post-training with a finite control model of each harness. This gives every checkpoint a fixed optimal policy to compare against, while retaining task instructions, parameterized actions, and controller feedback in the model's input.

We compare six harnesses, H0--H5, using H0 as the direct baseline.

\paragraph{Tasks and observations.}
We use three tasks from MiniWoB++~\citep{liu2018reinforcement}. In \texttt{click-checkboxes}, the agent toggles checkboxes to match the requested labels and submits the selection. In \texttt{click-checkboxes-soft}, the requested words and checkbox labels are matched by synonym meaning. In \texttt{ascending-numbers}, the agent clicks five tiles in numerical order, with the remaining numbers hidden after the first click. We retain the initially visible layout in the public observation, together with spatial tile IDs, the remaining tiles, and the number of completed clicks. Checkbox observations contain the instruction, each label and its checked status. All actor observations also expose the remaining decision budget and the harness's controller state.

The reward is one for exact task completion and zero otherwise, including partial completion. All explicit action and tool costs are zero. Each episode allows at most ten actor decisions, with discount $\gamma=0.95$ per decision. We compile the physical transitions, remaining budget, and controller memory into a finite acyclic graph. For every retained task, we verify a successful run in the native browser and separately compare the compiled transitions and termination rules with the page behavior. Training and exhaustive policy evaluation use these compiled graphs.

\paragraph{Harnesses.}
Table~\ref{tab:e1-harnesses} specifies the six interfaces. The program verifier approves an action when it avoids an incorrect terminal outcome, using the public instruction, page state, and the benchmark's fixed synonym vocabulary. Verification targets are checkbox submission and each of the five tile clicks. H4 uses a separate frozen Qwen3.5-2B reviewer whose prompt contains the public task and physical page state together with the proposed action, omitting the actor's remaining budget and controller memory. It greedily selects between single-token approve/reject codes. For H3 and H4, verification feedback becomes part of the next observation and is cleared by a native action. The H1 batch executes two ordered clicks within one actor decision and stops at native termination, so it receives only one discount step. All ordered pairs of click IDs are available, including repeated IDs.

\begin{table}[ht]
\centering\small
\caption{MiniWoB++ harnesses. Every menu includes the complete parameters of each action. We measure the common budget in actor decisions.}
\label{tab:e1-harnesses}
\begin{tabularx}{\linewidth}{p{0.55cm}p{2.2cm}Y}
\toprule
ID & Interface & Actions, execution, and controller state\\
\midrule
H0 & Direct & One native click or checkbox submission per decision.\\
H1 & Batched clicks & Adds every ordered pair of clicks as a single menu option, executed without an intervening actor call.\\
H2 & Automatic gate & Checks a commit action before execution. Rejection leaves the page unchanged, consumes the decision, and disables subsequent checks. Memory stores whether the gate is enabled.\\
H3 & Verify tool & Adds an optional program check of a proposed native action. Memory stores the checked action and its advisory verdict.\\
H4 & Reviewer tool & Uses H3's action menu and memory, with advice supplied by a frozen language-model reviewer.\\
H5 & Revised instructions & Uses H0's actions, observations, dynamics, and budget, with additional system instructions for matching labels, tracking tile IDs, correcting selections, and deciding when to submit.\\
\bottomrule
\end{tabularx}
\end{table}

\paragraph{Data splits.}
We generate tasks with the native pages and deduplicate them by full public task content before assigning splits. Table~\ref{tab:e1-splits} gives the fixed counts. Checkbox instances contain fewer than five boxes, and in the number task the first required number lies in one of the first three spatial positions. The 72-task training probe is fixed before training and is a subset of the training set. The six harnesses share the same task contents and splits. Test tasks are held out from parameter updates.

All MiniWoB++ experiments reported in this paper use the same fixed test set of 576 tasks generated from MiniWoB++, with 192 tasks per family. Calibration and training-probe tasks remain separate.

\begin{table}[ht]
\centering\small
\caption{MiniWoB++ task allocation. Splits have disjoint public contents, except that the training probe comes from training. Metrics average tasks within each family and then weight the families equally.}
\label{tab:e1-splits}
\begin{tabularx}{\linewidth}{lrrY}
\toprule
Split & Per family & Total & Use in post-training\\
\midrule
Training & 480 & 1,440 & Parameter updates\\
Calibration & 96 & 288 & Full learning curves\\
Training probe & 24 & 72 & Fixed training subset for learning curves\\
Test & 192 & 576 & Transfer evaluation\\
\bottomrule
\end{tabularx}
\end{table}

\paragraph{Policy readout and reference.}
The actor receives a system message specifying the task and harness rules and a JSON observation listing every available action. Each complete action, including its arguments, maps bijectively to one case-sensitive token from \texttt{A--Z/a--z}. We normalize the next-token logits over this full menu at sampling temperature one, using the model's chat template with thinking disabled. This categorical distribution defines $\Pi_{\theta,\I_h}$ throughout training and evaluation. The base model and tokenizer are Qwen3.5-2B. An independent frozen copy, without the trained adapter, reads the same messages to define $\piref_{\I_h}$ at each state. The reference stays fixed across checkpoints. Because H5 changes the reference model's input as well as the actor's, we solve its regularized optimum separately from H0's even though their execution dynamics coincide.

\begin{table}[t]
\centering\small
\caption{Post-training settings. All harnesses share the same update and task-draw budgets, with computation depending on the number of reachable states and menu entries.}
\label{tab:e1-optimization}
\begin{tabularx}{\linewidth}{p{3.5cm}Y}
\toprule
Parameter & Setting\\
\midrule
Base / reference / H4 reviewer & Qwen3.5-2B with frozen base, reference, and reviewer weights\\
Trainable adapter & LoRA rank 16, scale $\alpha=32$, dropout 0, no bias updates\\
LoRA projection targets & Attention: \texttt{q/k/v/o\_proj}\newline Linear attention: \texttt{in\_proj\_qkv}, \texttt{in\_proj\_z}, \texttt{in\_proj\_b}, \texttt{in\_proj\_a}, \texttt{out\_proj}\newline MLP: \texttt{gate/up/down\_proj}\\
Optimizer & AdamW with a constant learning rate $10^{-5}$, $\beta=(0.9,0.999)$, $\epsilon=10^{-8}$, weight decay 0, gradient norm clipping at 1\\
Training budget & 1,024 updates per harness at four tasks per update (4,096 task draws)\\
Checkpoint schedule & Initialization and every 64 updates (17 checkpoints at the full budget)\\
Numerical precision & BF16 for model forward passes\newline FP32 for trainable parameters and optimizer states\newline FP64 for categorical log-softmax and dynamic programming\\
State microbatch & Eight states grouped by prompt length and menu size\\
Prompt length & At most 2,048 tokens without truncation\\
\bottomrule
\end{tabularx}
\end{table}

\paragraph{Training distribution and objective.}
Each harness starts from the same base weights and initial LoRA adapter~\citep{hu2022lora}. To retain correction states as successful behavior becomes more frequent, training uses
\begin{equation}
\mu_h=\tfrac12\delta_{s_{0,h}}+\tfrac12\nu_h,
\qquad
J_{\mu_h}(\theta)=\E_{\mu_h,\Pi_{\theta,\I_h},P_h}
\sum_{t\ge0}\gamma^t\left[r_h(s_t,a_t)-\tau\log\frac{\Pi_{\theta,\I_h}(a_t\mid s_t)}{\piref_{\I_h}(a_t\mid s_t)}\right],
\label{eq:e1-objective}
\end{equation}
with $\tau=0.1$. Here $s_{0,h}$ is the native task start. We form $\nu_h$ from all other reachable nonterminal states, stratifying by correction category or completion progress, remaining-budget bin, and controller memory. Checkbox categories are correct, missing selections only, extra selections only, and both missing and extra selections. For the number task, categories are defined by the number of completed clicks. Budget bins are 1--3, 4--6, and 7--9 remaining decisions. We assign equal mass to nonempty strata and then to states within each stratum. Each start has a legal action prefix from the native state, and the objective begins at that start with its remaining budget. This gives every reachable decision state positive training mass while leaving the statewise dynamics, rewards, and optimal action preferences fixed.

\paragraph{Exact-PG updates.}
Each update samples four tasks with replacement, first choosing a task family uniformly and then an instance uniformly within that family. For each task, dynamic programming evaluates the current policy's regularized value and advantage, and propagates discounted visitation from $\mu_h$. Exact-PG sums the policy gradient over this task's reachable states and complete action menu. Gradients for the four tasks are evaluated at the same parameters and averaged once per optimizer step. The objective thus weights tasks equally regardless of graph size. A separate soft Bellman solve provides $\pistar_{\I_h}$ and $F^*_{\I_h}$ as evaluation targets. Table~\ref{tab:e1-optimization} gives the optimization settings.

Each training run uses one RTX 4090 GPU.

\paragraph{Geometry and return measurements.}
At initialization and each saved checkpoint, we read the complete action distribution at every reachable nonterminal state of all calibration and training-probe tasks. For a task $u$, let $\Sset_{h,u}$ denote this fixed state bank. Its geometry error is
\begin{equation}
\operatorname{MSE}_{h,u}(\theta)
=\frac{1}{|\Sset_{h,u}|}\sum_{s\in\Sset_{h,u}}
\frac{\left\lVert\tau\proj\!\left[\log\Pi_{\theta,\I_h}(\cdot\mid s)-\log\pistar_{\I_h}(\cdot\mid s)\right]\right\rVert_2^2}{|\Aset_{h,u}(s)|}.
\label{eq:e1-mse}
\end{equation}
We average this quantity within task families and then across the three families. The fixed bank makes the trajectory of this error comparable across checkpoints, including decisions that become rare under the learned policy. Alongside it, we compute native-start success probability, $J$, $C$, and the squared score error weighted by the evaluated policy's discounted occupancy. Success is the undiscounted probability of exact completion, evaluated by dynamic programming. Mismatch is computed independently as $\tau/(1-\gamma)$ times the occupancy-weighted $\KL(\Pi_{\theta,\I_h}\Vert\pistar_{\I_h})$, allowing comparison with $C-J$. We sum over nonterminal states without renormalizing their occupancy mass.

\paragraph{Transfer and uncertainty.}
For transfer, we evaluate the six trained actors at the common final update 1,024 under all six deployment harnesses on the fixed test set, which gives the $6\times6$ matrix in Figure~\ref{fig:training-geometry}. Every cell uses its deployment harness's state graph, menu, frozen reference, and optimal target. Comparisons within a column therefore hold the deployment control problem fixed.

The learning curves in Figure~\ref{fig:training-geometry} use all 288 calibration tasks at initialization and every 64 updates through update 1,024. For both success and geometry MSE, we form pointwise 95\% percentile intervals from 2,000 bootstrap resamples of tasks within each family. The same task draws are shared across metrics and checkpoints, with equal weight on each family. Changes from initialization use paired differences under these same draws. All measured checkpoints are plotted without smoothing. We make the same measurements on the 72-task training probe. From initialization to update 1,024, the geometry error falls by 69\% to 90\% across the six harnesses, both on the training probe and on the calibration tasks, which play no role in parameter updates. For each harness, the two reductions differ by less than one percentage point. Post-training thus improves the geometry on held-out and training tasks alike.

\subsection{Readout Study}
\label{app:exp-readout}
\paragraph{Tasks and frozen checkpoint.}
The main-text comparison uses the H0-trained Qwen3.5-2B checkpoint at post-training update 1,024. Its weights remain frozen throughout evaluation. We evaluate it on the fixed test set in Table~\ref{tab:e1-splits}, the same 576 tasks used by the adaptation and prediction studies below. R0 therefore reproduces the source checkpoint's H0 results in Table~\ref{tab:adaptation-endpoints}.

\paragraph{Presentation variants.}
R0 uses the canonical H0 system message, JSON observation, and action-code mapping. R1 reverses the order of top-level observation fields while preserving their contents and the nested action dictionary. R2 replaces the action dictionary with a list of records containing the same code and action fields. R3 shifts the action-code assignment cyclically by one position, R4 reverses it, and R5 uses a fixed random permutation for each menu size. These three variants preserve the order and full arguments of the semantic actions and update the displayed code assignment and decoder together. We align every probability table by semantic action before comparison. The system message and all task information are unchanged, and inputs are evaluated without truncation.

\paragraph{Shared reference and exact evaluation.}
We fix the semantic reference $q_0(a\mid s)$ to the frozen base model's distribution under R0 and assign this same probability to the corresponding code in each variant. The evaluated actor reads each variant normally. All conditions use H0's native actions and dynamics, ten model decisions, $\gamma=0.95$, $\tau=0.1$, and zero explicit costs. The full legal single-token menu is evaluated at sampling temperature one. Model forward passes use BF16 with state microbatches of eight, followed by FP64 probability normalization and control calculations. We enumerate reachable states, solve the optimum separately in each code ordering, and compute return and success from the measured semantic policy by dynamic programming. Mismatch is computed independently from occupancy-weighted $\KL(\Pi\|\pistar)$. The largest ceiling difference is $2.2\times10^{-16}$ and the largest optimal-action probability difference is $8.3\times10^{-16}$ across variants.

\paragraph{Measurements and uncertainty.}
Choice change is the fraction of states where the most probable semantic action differs from R0. Readout RMS is the square root of the mean squared entries of $\eta=\tau\proj(\log\Pi_{\theta,R_v}-\log\Pi_{\theta,R_0})$. We average state-level measurements within each task, then average tasks within each family and give the three families equal weight. All variants are evaluated on the same tasks. We form percentile intervals from 2,000 paired task bootstrap resamples within families. These intervals quantify task variation at the fixed checkpoint. Table~\ref{tab:readout-full} reports the readout and performance measurements.

\begin{table}[htbp]
\centering\small
\caption{H0 readout results, with 95\% paired-task bootstrap intervals below each mean. RMS intervals apply the square root to the mean-squared drift interval endpoints. The target drift is numerically zero in every presentation.}
\label{tab:readout-full}
\setlength{\tabcolsep}{3pt}
\begin{tabular}{@{}lcccc@{}}
\toprule
Variant & Readout RMS & Mismatch $M$ & Return $J$ & Success (\%)\\
\midrule
R0 & 0.000 & 0.015 & $0.423$ & 85.0\\
   & [0.000, 0.000] & [0.010, 0.021] & [$0.409$, $0.436$] & [84.0, 86.1]\\
\addlinespace[2pt]
R1 & 0.200 & 0.484 & $-0.046$ & 19.2\\
   & [0.198, 0.203] & [0.460, 0.507] & [$-0.072$, $-0.020$] & [17.0, 21.4]\\
\addlinespace[2pt]
R2 & 0.054 & 0.023 & $0.415$ & 86.0\\
   & [0.053, 0.054] & [0.018, 0.028] & [$0.401$, $0.428$] & [85.1, 86.9]\\
\addlinespace[2pt]
R3 & 0.181 & 0.148 & $0.289$ & 57.1\\
   & [0.179, 0.183] & [0.143, 0.155] & [$0.274$, $0.303$] & [55.9, 58.3]\\
\addlinespace[2pt]
R4 & 0.181 & 0.130 & $0.308$ & 68.2\\
   & [0.179, 0.182] & [0.124, 0.136] & [$0.292$, $0.323$] & [66.9, 69.5]\\
\addlinespace[2pt]
R5 & 0.168 & 0.110 & $0.328$ & 67.6\\
   & [0.166, 0.170] & [0.104, 0.116] & [$0.312$, $0.342$] & [66.2, 68.9]\\
\bottomrule
\end{tabular}
\end{table}

\subsection{Adaptation After an Interface Revision}
\label{app:exp-adaptation}
\paragraph{Revision and source checkpoint.}
We study H0-to-H1 adaptation using the model, task families, action decoder, and harness definitions of Appendix~\ref{app:e1}. H1 retains H0's native actions and adds ordered pairs of clicks executed within one actor decision, stopping at native termination. The source checkpoint $\theta_0$ is the H0-trained Qwen3.5-2B model at post-training update 1,024.

\paragraph{Tasks and splits.}
Training reuses the 1,440 training tasks and the coverage mixture $\mu_h$ in Eq.~\eqref{eq:e1-objective}. The evaluation reuses all 576 tasks in Table~\ref{tab:e1-splits}, with 192 instances per family. These tasks are held out from parameter updates and checkpoint selection. We reuse the same task graphs, action menus, and reference probabilities as the post-training study. All deployment metrics start from the native task starts $\rho_0$.

\paragraph{Training branches and budget.}
Both branches inherit the same source LoRA weights and reset AdamW's first and second moments. One continues under H0, giving $\theta_s$, and the other adapts under H1, giving $\theta_t$. Both use paired task draws. Each receives 1,024 additional Exact-PG updates with four tasks per update, for 4,096 task draws. We retain the post-training settings in Table~\ref{tab:e1-optimization}, including LoRA rank 16 and scale 32, constant learning rate $10^{-5}$, BF16 forward passes, and state microbatches of eight. Each interface's dynamics, costs, frozen base-model reference, $\gamma=0.95$, and $\tau=0.1$ stay fixed throughout training and evaluation.

\paragraph{Evaluation schedule and measurements.}
We evaluate at additional updates 0, 32, 64, 128, 256, 512, and 1,024. The source checkpoint is evaluated under both interfaces, and each later checkpoint from both branches is evaluated under both interfaces, giving 26 complete evaluations. The budget and checkpoint schedule were fixed before adaptation. The endpoint comparisons use update 1,024, and the curves include every scheduled checkpoint.

The oracle solves the soft Bellman equations for $C_j$ and $\pistar_{\I_j}$. Complete action probabilities and the evaluated policy's forward discounted visitation determine $M$ through Eq.~\eqref{eq:gap}. Terminal states contribute zero loss, and nonterminal visitation mass is retained without renormalization. Independently, Bellman policy evaluation computes $J$, while undiscounted policy evaluation gives task success. The residual in $J=C-M$ checks numerical consistency. These measurements use the compiled task graphs and full action distributions.

\paragraph{Comparisons.}
For each branch, we compare the target-interface mismatch with the fixed recovery threshold $C_2-J(\theta_0,\I_1)$. We separately evaluate recovery beyond the source pair, interface ranking for the updated checkpoint, and the return difference between the H1-adapted pair and continued training under H0. Evaluating both branches under H1 also measures the effect of the training interface at a common deployment interface. We report the component gains $A_1,A_2$ and their difference $\Omega$ from Appendix~\ref{app:adaptation}, alongside regularized return and task success.

\paragraph{Uncertainty and computation.}
Metrics average tasks within each family and weight the families equally. We form pointwise 95\% percentile intervals from 2,000 paired bootstrap resamples within families. The same resampling indices are shared across checkpoints, interfaces, metrics, and paired differences. Each branch trains on one RTX 4090 GPU, and a third GPU evaluates checkpoints.

\paragraph{Adaptation results.}
H1 has a higher ceiling than H0, and both stay fixed across checkpoints. The frozen swap increases mismatch enough to outweigh this gain and lower return. H1 adaptation then reduces mismatch below the recovery threshold, and the final adapted pair exceeds both the source H0 pair and H0 continuation under H0. Table~\ref{tab:adaptation-endpoints} gives the measurements of the source checkpoint and both final checkpoints. Rising return under H1 and falling return under H0 together reverse the ranking along the adaptation branch.

The interface ranking reverses before the adapted pair recovers the source pair's return. At the first recorded checkpoint with a positive mean recovery difference, its paired interval still includes zero. By the final budget, the paired intervals for recovery and improvement over H0 continuation both lie above zero. Table~\ref{tab:adaptation-comparisons} reports these endpoint comparisons, and Figure~\ref{fig:diagnostic-adaptation} shows the recovery and ranking trajectories. In the notation of Appendix~\ref{app:adaptation}, $\I_1$ is H0 and $\I_2$ is H1.

\begin{table}[htbp]
\centering\small
\caption{Deployment outcomes of the source checkpoint and both final checkpoints. Both training branches receive 1,024 additional updates. $J$ is regularized return, $M$ is occupancy-weighted decision loss, and success is the undiscounted probability of completing the task.}
\label{tab:adaptation-endpoints}
\begin{tabular}{llrrr}
\toprule
Checkpoint & Deployment & $J$ & $M$ & Success (\%)\\
\midrule
Source & H0 & 0.42289 & 0.01485 & 85.05\\
Source & H1 & 0.31285 & 0.13737 & 72.90\\
H0 continuation & H0 & 0.42294 & 0.01481 & 83.37\\
H0 continuation & H1 & 0.30404 & 0.14619 & 75.66\\
H1 adaptation & H0 & 0.38642 & 0.05132 & 78.31\\
H1 adaptation & H1 & 0.43133 & 0.01889 & 88.17\\
\bottomrule
\end{tabular}
\end{table}

\begin{table}[htbp]
\centering\small
\caption{Paired comparisons with 95\% task bootstrap intervals. Final comparisons use the H1-adapted checkpoint at 1,024 additional updates. Recovery compares it under H1 with the source H0 pair. Interface ranking compares H1 and H0 at the same adapted checkpoint. Gain over H0 continuation compares the two branches at the same update budget, each under its training harness.}
\label{tab:adaptation-comparisons}
\begin{tabularx}{\linewidth}{@{}Yr@{}}
\toprule
Quantity & Mean [95\% CI]\\
\midrule
Ceiling gain $C_2-C_1$ & $0.01248\ [0.00698,\,0.01754]$\\
Frozen mismatch increase & $0.12252\ [0.11487,\,0.12961]$\\
Frozen return change & $-0.11004\ [-0.11558,\,-0.10438]$\\
Recovery threshold $T$ & $0.02733\ [0.01984,\,0.03520]$\\
\midrule
Final recovery difference & $0.00844\ [0.00135,\,0.01447]$\\
Final interface ranking & $0.04491\ [0.03653,\,0.05254]$\\
Final gain over H0 continuation & $0.00840\ [0.00254,\,0.01411]$\\
$A_1$: change under H0 & $-0.03647\ [-0.04033,\,-0.03295]$\\
$A_2$: change under H1 & $0.11848\ [0.10964,\,0.12658]$\\
$\Omega=A_2-A_1$ & $0.15495\ [0.14430,\,0.16493]$\\
\bottomrule
\end{tabularx}
\end{table}

\subsection{Adaptation Across Harness Swaps}
\label{app:exp-swaps}
Among the 30 ordered swaps between H0--H5, a frozen comparison rejects 22, and 13 of those have $T\le0$, which leaves nine. Each of the nine starts from the checkpoint post-trained under its source harness in Appendix~\ref{app:e1} and adapts under the target harness with the protocol of Appendix~\ref{app:exp-adaptation}. We recorded the predictions for the eight swaps other than H0$\to$H1 before adapting them.

Table~\ref{tab:swaps} lists the nine swaps. For the eight swaps recorded in advance, $T-r_2$ orders the gains with Spearman $\rho=0.95$, the ceiling gain alone with $\rho=0.86$, and the frozen return change with $\rho=0.17$. Subtracting the frozen return change and the gain from $T$ gives the target mismatch before and after adaptation. It falls from between 0.015 and 0.146 to at most 0.019. The observed gain exceeds $T-r_2$ in every swap, by 0.004 on average. The ranking predicted from the checkpoint post-trained under the target has the correct sign in all eight recorded swaps and orders the observed rankings with Spearman $\rho=0.91$. After adapting from H1 to H0, the checkpoint prefers H0 by 0.041 and still falls short of the source pair, so a ranking reversal alone does not show a net gain. Adapting from H0 to H5 also beats continued training under H0 by $0.085\ [0.078,\,0.092]$. The 13 swaps with $T\le0$ include all five swaps from H5, with $T$ between $-0.136$ and $-0.016$.

\begin{table}[htbp]
\centering\small
\caption{The nine swaps that a frozen comparison rejects and that have $T>0$, ordered by the observed gain. The ceiling gain $\Delta C$, the frozen return change, $T$, and the predicted gain $T-r_2$ come from the post-trained checkpoints before adaptation. The gain is the adapted pair's return minus the source pair's return after 1,024 updates. Ranking is the adapted checkpoint's return under the target minus its return under the source. Success is the change in task success from the source pair, in percentage points. The prediction for H0$\to$H1 was computed after its adaptation.}
\label{tab:swaps}
\setlength{\tabcolsep}{4pt}
\begin{tabular}{@{}lrrrrrrr@{}}
\toprule
Swap & $\Delta C$ & Frozen & $T$ & $T-r_2$ & Gain & Ranking & Success\\
\midrule
H1$\to$H5 & $0.068$ & $-0.054$ & $0.092$ & $0.081$ & $0.085$ & $0.167$ & $+9.0$\\
H0$\to$H5 & $0.081$ & $-0.004$ & $0.095$ & $0.085$ & $0.085$ & $0.111$ & $+8.7$\\
H3$\to$H1 & $0.078$ & $-0.040$ & $0.099$ & $0.075$ & $0.083$ & $0.169$ & $+11.7$\\
H4$\to$H1 & $0.075$ & $-0.040$ & $0.089$ & $0.066$ & $0.071$ & $0.164$ & $+12.6$\\
H2$\to$H5 & $0.038$ & $-0.018$ & $0.053$ & $0.042$ & $0.045$ & $0.084$ & $-0.5$\\
H1$\to$H2 & $0.030$ & $-0.055$ & $0.054$ & $0.039$ & $0.042$ & $0.082$ & $+9.8$\\
H0$\to$H1 & $0.012$ & $-0.110$ & $0.027$ & $0.003$ & $0.008$ & $0.045$ & $+3.1$\\
H4$\to$H3 & $-0.003$ & $-0.004$ & $0.011$ & $-0.009$ & $-0.001$ & $-0.001$ & $+0.1$\\
H1$\to$H0 & $-0.012$ & $-0.060$ & $0.011$ & $-0.003$ & $-0.002$ & $0.041$ & $-2.5$\\
\bottomrule
\end{tabular}
\end{table}

\subsection{Calibration of Finite Mismatch Measurements}
\label{app:exp-measurement}
\paragraph{Estimator.}
For each task and deployed pair, draw $m$ independent pilot trajectories. For each trajectory, draw $T$ with $\Pr(T=t)=(1-\gamma)\gamma^t$ and retain its state $s_T$. Retaining one state per independent trajectory gives independent occupancy samples. A draw after termination contributes zero to the average over all $m$ draws. At each nonterminal sampled state with $K$ actions, draw $N$ independent actions and form
\begin{equation}
\widetilde p(a\mid s)=\frac{n_a+\alpha}{N+K\alpha}.
\label{eq:smoothed-action-probability}
\end{equation}
These smoothed distributions enter the mismatch estimate in Eq.~\eqref{eq:sampled-mismatch}. Each nonterminal state draw uses the complete deployed action menu. Pilot and action-sampling queries are both counted.

\paragraph{Candidate budgets and smoothing.}
The grid uses $m\in\{64,256\}$ and $N\in\{16,64,256,1024\}$. Four smoothing rules use a pseudocount of $0.1$ or $0.5$ per action, or a total pseudocount $\beta\in\{0.5,1\}$ divided across the $K$ actions. This gives 32 candidates. Calibration uses 32 independent measurement pools, with shared nested prefixes across candidates. For each candidate, the calibration scorer computes the expected absolute-mismatch bias using exact occupancy and the binomial marginal distribution of each action count. The predictor receives only the sampled observations.

\paragraph{Selection criteria.}
A candidate must meet three criteria over H0 and H1 at every adaptation node: absolute expected mismatch bias at most $0.005$ averaged over families, bias at most $0.01$ within every family, and aggregate return-difference RMSE at most $0.01$ for recovery and ranking across measurement pools. It must also reach RMSE at most $0.03$ on the configuration calibration set. Among passing candidates, we select the lowest mean realized actor-query cost. Costs within 1\% tie and are resolved by worst-node difference RMSE, $m$, $N$, and the fixed smoothing-rule order.

\begin{figure}[htbp]
\centering
\includegraphics[width=\linewidth]{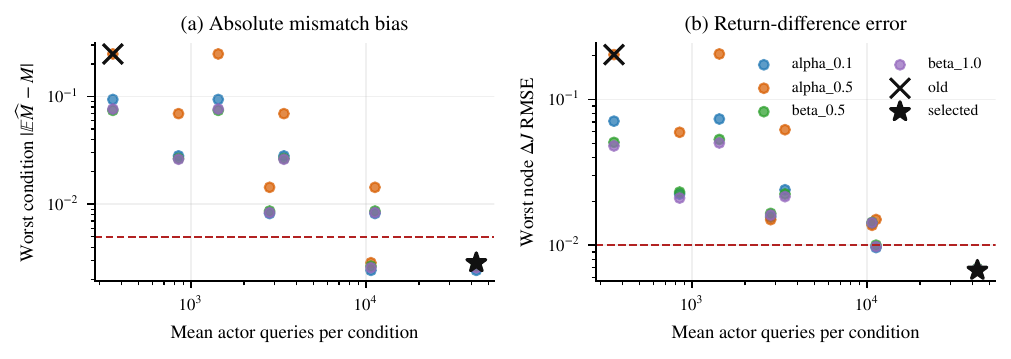}
\caption{Measurement accuracy depends on the query budget and smoothing rule. Each point is a calibration candidate. (a) Largest absolute mismatch bias across deployment conditions. (b) Largest aggregate return-difference RMSE across adaptation nodes. Dashed lines mark calibration thresholds. The cross marks the initial budget and the star marks the selected estimator.}
\label{fig:measurement-budget}
\end{figure}

Four candidates pass, all with $m=256$ and $N=1024$. The selected rule uses $\alpha=0.5$ per action and has configuration calibration RMSE $0.00887$. Each test prediction uses a single measurement pool. Table~\ref{tab:measurement-calibration} compares the initial and selected budgets, and Figure~\ref{fig:measurement-budget} shows the full sweep.

\begin{table}[htbp]
\centering\small
\caption{Calibration of the measurement budget. Both columns use a pseudocount of $0.5$ per action. Bias refers to the magnitude of the expected mismatch estimation error. Query counts include pilot actions and frequency samples at nonterminal states.}
\label{tab:measurement-calibration}
\begin{tabularx}{\linewidth}{Yrr}
\toprule
Calibration quantity & $m=64$, $N=16$ & $m=256$, $N=1024$\\
\midrule
Largest condition-level absolute bias in $M$ & 0.24805 & 0.00287\\
Largest family-level absolute bias in $M$ & 0.40320 & 0.00442\\
Worst-node return-difference RMSE & 0.20259 & 0.00672\\
Mean actor queries per task and condition & 356.9 & 42,553.3\\
\bottomrule
\end{tabularx}
\end{table}

\paragraph{Why the budget changes the estimate.}
Eq.~\eqref{eq:smoothed-action-probability} mixes the empirical action distribution with a uniform distribution of weight $K\alpha/(N+K\alpha)$. For a 30-action menu with $\alpha=0.5$, this weight is 48.4\% at $N=16$ and 1.44\% at $N=1024$. Larger action budgets reduce this distortion, while more occupancy samples stabilize the aggregate estimate. On the test configurations of Table~\ref{tab:diagnostic-prediction}, increasing the budget reduces prediction RMSE from $0.01132$ to $0.00230$. The selected budget uses about 119 times as many actor queries on calibration.

\subsection{Independent Prediction of Deployment Outcomes}
\label{app:exp-prediction}
\paragraph{Tasks and checkpoints.}
The prediction study uses the fixed test set in Table~\ref{tab:e1-splits}, with 576 tasks divided equally across the three MiniWoB++ families. Estimator calibration uses a separate set of 72 tasks, 24 per family, generated in the same way as the splits in Table~\ref{tab:e1-splits} and disjoint from all of them. This set is fixed before estimator selection and plays no role in the reported test results. We use the H0 checkpoint at post-training update 1,024 and the H1-adapted checkpoints from Appendix~\ref{app:exp-adaptation}. All weights are frozen for this study. The control objective uses $\gamma=0.95$, $\tau=0.1$, and a horizon of ten model decisions.

\paragraph{Execution and encoding sweep.}
Both final checkpoints are deployed through H0's native action menu. We cross execution settings with three encodings: the canonical presentation R0, reversed top-level observation fields R1, and a cyclic permutation of action codes R3. The encodings preserve the semantic actions and their execution. The reference is the frozen base model's distribution under R0, aligned to the same semantic actions in all three presentations. Prompts are held fixed across execution settings and omit numerical reliability and cost values.

The baseline has click reliability $p=0.8$, click cost $c=0.01$, and encoding R0. One sweep varies $p\in\{0.5,0.6,0.7,0.75,0.8,0.85,0.9,0.95,1\}$ at fixed $c=0.01$. A second varies $c\in\{0,0.0025,0.005,0.0075,0.01,0.015,0.02,0.03,0.05\}$ at fixed $p=0.8$. Failed clicks leave the physical page unchanged and consume the decision and its cost. Submission retains the native task semantics. Calibration uses $p\in\{0.5,0.8,1\}$ and $c\in\{0,0.01,0.05\}$ in their respective sweeps. Crossing these settings with the encodings gives 15 calibration and 36 test configurations per checkpoint. The two checkpoints yield 72 test comparisons with their respective baselines.

\paragraph{Prediction and scoring.}
For each task and configuration, an oracle solves the control problem for $C$ and the queried optimal action distributions. The predictor obtains independent pilot trajectories and finite categorical samples from the deployed policy. Eq.~\eqref{eq:sampled-mismatch} gives $\widehat M$ for the candidate and baseline. Their predicted return difference is
\begin{equation}
\widehat{\Delta J}=\Delta C-\Delta\widehat M.
\label{eq:predicted-return}
\end{equation}
Appendix~\ref{app:exp-measurement} specifies the estimator and calibration. All test predictions are saved before scoring starts. The scoring process computes exact $J$ and occupancy-weighted KL from full policy probabilities. The predictor's data access is limited to samples, pilot states, the reference, oracle ceilings, and queried optimal targets.

\paragraph{Adaptation comparisons.}
We evaluate the saved H1-adapted checkpoints at 0, 32, 64, 128, 256, 512, and 1,024 updates under native H0 and H1. Each interface retains its frozen base-model reference. At every node, the predictor estimates the recovery difference $J(\theta_t,\mathrm{H1})-J(\theta_0,\mathrm{H0})$ and the ranking difference $J(\theta_t,\mathrm{H1})-J(\theta_t,\mathrm{H0})$. The source H0 baseline receives an independent measurement whose cost is included.

\paragraph{Baselines and access.}
Ceiling-only prediction sets $\Delta\widehat M=0$. The frequency rollout baseline estimates regularized returns directly, using repeated queries at each visited state to estimate the selected action's log probability. Calibration selects 64 such samples per decision with additive smoothing $\alpha=0.5$. This baseline receives the same actor-query allowance as the diagnostic, including its pilot queries and baseline measurement. Reference computation and oracle construction are recorded separately from actor queries.

\begin{table}[htbp]
\centering\small
\caption{Prediction error on the fixed test tasks. Configuration RMSE averages over 72 checkpoint and configuration comparisons. Recovery and ranking RMSE average over seven saved adaptation checkpoints. No estimator reads exact action probabilities.}
\label{tab:diagnostic-prediction}
\begin{tabularx}{\linewidth}{Yrrr}
\toprule
Estimator & Configuration RMSE & Recovery RMSE & Ranking RMSE\\
\midrule
Ceiling minus measured mismatch & 0.00230 & 0.00245 & 0.00286\\
Ceiling alone & 0.23470 & 0.05744 & 0.05342\\
Frequency rollout, same query budget & 0.00240 & 0.02672 & 0.02667\\
\bottomrule
\end{tabularx}
\end{table}

\begin{figure}[htbp]
\centering
\includegraphics[width=\linewidth]{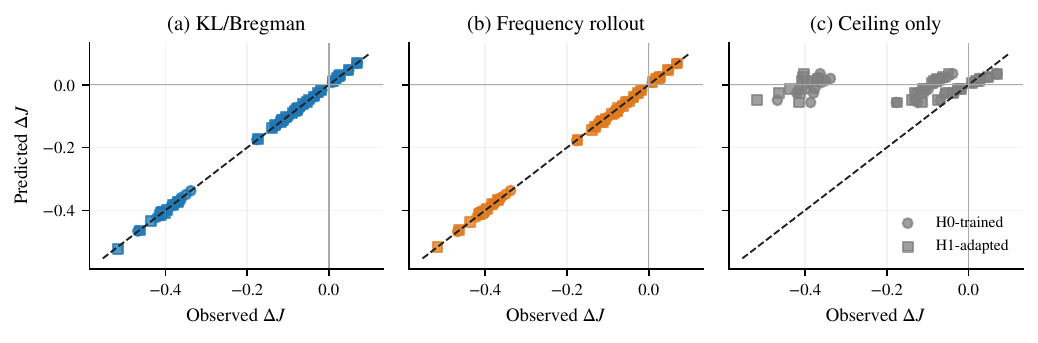}
\caption{Measured mismatch predicts the return realized under a harness revision. Each point compares a candidate with its baseline on the fixed test tasks. The diagonal marks agreement between predicted and observed return changes. Circles and squares denote the H0-trained and H1-adapted checkpoints. (a) Ceiling minus measured mismatch. (b) Direct rollouts with action probabilities estimated from frequencies at the same actor-query budget. (c) Ceiling alone.}
\label{fig:finite-diagnostic-prediction}
\end{figure}

\paragraph{Prediction results.}
Figure~\ref{fig:finite-diagnostic-prediction} compares predicted and observed return changes across the configuration sweep.
Table~\ref{tab:diagnostic-prediction} reports errors in aggregate return differences, where each comparison first averages returns equally across task families. In the configuration sweep, the mismatch-based prediction identifies the direction of all 70 changes outside a $0.01$ tie region and orders all 1,195 configuration pairs outside that region within each checkpoint. Classifying changes as harmful, tied, or beneficial gives 71 correct predictions out of 72.

In the adaptation study, at the final update 1,024, the exact recovery and ranking differences are $0.00844$ and $0.04491$, with intervals in Table~\ref{tab:adaptation-comparisons}. Their predictions are $0.00579$ with 95\% interval $[-0.00116,0.01228]$ and $0.04195$ with interval $[0.03371,0.05015]$. Using the same $0.01$ tie region, three-class predictions agree with the exact values at six of seven recovery nodes and all seven ranking nodes. Figure~\ref{fig:diagnostic-adaptation} in Section~\ref{sec:exp-prediction} shows the predicted and observed trajectories.

\paragraph{Uncertainty and resources.}
We average within each family and give equal weight to the three families. Intervals use 2,000 paired task bootstrap resamples within families. They describe task-sampling uncertainty for the fixed checkpoints. Finite action samples are simulated from cached distributions of the actual language-model checkpoints. The new probability measurements took 2.19 GPU-hours.

\subsection{Coding-Agent Harness Study}
\label{app:exp-realagents}
\paragraph{Harnesses and benchmark.}
The study uses the native configurations of DeepAgents, Pi coding-agent, DeepSeek Harness, Terminus-2, and mini-SWE-agent. All file and command tools run inside the benchmark task container. The framework owns its loop, tool feedback, delegation where configured, compression, retries, and termination. Terminus-2 retains tmux, completion confirmation, and summarization. Mini-SWE-agent uses its official tool-call profile. Terminal-Bench Lite contains 39 tasks, scored by the official verifier. The agent never sees scoring data or gold patches.

\paragraph{Accuracy.}
Each harness has three independent runs. Accuracy is the mean of the three single-run success rates over all planned tasks, and the reported spread is the sample standard deviation across runs. Timeouts and failures remain in the denominator as unresolved tasks.

\paragraph{Models and budgets.}
The target is Qwen3.5-9B with thinking enabled, temperature 0.6, an output limit of 32,768 tokens, and a context window of 65,536 tokens. The teacher is GPT-5.6-luna with the same output and context limits and high reasoning effort. Each trajectory is capped at 50 top-level decisions and uses the task's own time limit.

\paragraph{Recorded states and sampling.}
At each model call, we record the request context, tool definitions, controller state, and remaining budgets. Summary, subagent, and retry calls count as decisions. To sample a decision, we resend the recorded request and decode each sample with the harness's parser, without executing tools. Samples measure the next action distribution at the recorded state.

\paragraph{Probe tasks and trajectories.}
The probes use three Terminal-Bench Lite tasks: \texttt{acl-permissions-inheritance}, \texttt{analyze-access-logs}, and \texttt{bank-trans-filter}. For each harness, the target and the teacher each contribute three trajectories per task, taken from separate runs of their own. A trajectory enters an estimate only when every recorded state in it has a complete probe. The teacher estimates use 7 of 9 trajectories for Pi coding-agent and all 9 for the other harnesses. The target estimates use 7, 6, and 8 of 9 trajectories for Pi coding-agent, DeepAgents, and DeepSeek Harness, and all 9 for Terminus-2 and mini-SWE-agent.

\paragraph{Common action categories.}
Each decoded output maps to one of 13 categories: inspect, modify, verify, setup, run, plan, delegate, wait, stop, context update, return, opaque, and invalid. File reads, listings, and searches map to inspect, and writes, edits, and deletions to modify. Todo and planning operations map to plan, and subagent calls to delegate. Shell commands, including Terminus-2 keystrokes at a shell prompt, are parsed, and a command maps to its dominant operation. Commands whose program the parser cannot resolve map to opaque, and responses without a countable native action map to invalid. The categories merge actions with different arguments and effects, so the estimates compare decisions at the level of these categories.

\begin{figure}[htbp]
\centering
\includegraphics[width=0.55\linewidth,trim=0 12 184 0,clip]{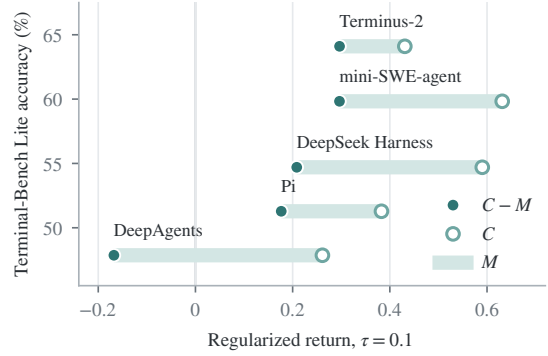}
\caption{Ceiling and mismatch of five coding-agent harnesses for a frozen Qwen3.5-9B. Estimates of Table~\ref{tab:real-agents} on three Terminal-Bench Lite probe tasks with $\tau=0.1$. Each harness sits at its accuracy over all 39 tasks. The open circle marks $\widetilde C$, the filled circle marks $\widetilde C-\widetilde M$, and the bar between them has length $\widetilde M$.}
\label{fig:real-agents-a}
\end{figure}

\paragraph{Estimators.}
At a recorded state $s$ under harness $h$, let $p_h(\cdot\mid s)$ and $q_h(\cdot\mid s)$ be the target's and the teacher's category distributions, estimated from 128 and 64 samples with a pseudocount of $0.5$ per category. The objective uses the reference $p_h$, a binary terminal reward $Y$, and no discount. The ceiling estimate averages the teacher's regularized return over its complete trajectories $s_0,\ldots,s_{L-1}$:
\begin{equation}
\widetilde C_h=\E_{q_h}\Bigl[Y-\tau\sum_{t=0}^{L-1}\KL\bigl(q_h(\cdot\mid s_t)\,\big\|\,p_h(\cdot\mid s_t)\bigr)\Bigr].
\label{eq:teacher-ceiling}
\end{equation}
The mismatch estimate applies Proposition~\ref{thm:probe-gap} with $q_h$ in place of the optimum. Without discounting, the factor $1/(1-\gamma)$ becomes a sum over the decisions of each complete target trajectory:
\begin{equation}
\widetilde M_h=\frac{1}{4\tau}\,\E_{p_h}\Bigl[\sum_{t=0}^{L-1}\Bigl\|\tau\proj\log\frac{p_h(\cdot\mid s_t)}{q_h(\cdot\mid s_t)}\Bigr\|_2^2\Bigr].
\label{eq:teacher-mismatch}
\end{equation}
We use $\tau=0.1$. With $p_h$ as the reference, the target's regularized return equals its success probability, so the exact ceiling minus mismatch also equals it. The estimates put a teacher that need not be optimal in place of the optimum, so $\widetilde C_h$ estimates a lower bound on the ceiling and $\widetilde M_h$ measures distance to the teacher. We therefore compare the ordering of $\widetilde C_h-\widetilde M_h$ with the ordering of accuracy. The mismatch per decision in Figure~\ref{fig:real-agents} is $\widetilde M_h$ divided by the mean number of decisions per complete target trajectory. Table~\ref{tab:real-agents-full} lists the three estimates and the decisions per trajectory, and Figure~\ref{fig:real-agents-a} plots the estimates against accuracy.

\begin{table}[tb]
\centering\small
\caption{Terminal-Bench Lite probe results at $\tau=0.1$. Decisions are the mean number of model calls per complete target trajectory.}
\label{tab:real-agents-full}
\begin{tabular}{@{}lrrrr@{}}
\toprule
Harness & Decisions & $\widetilde C_h$ & $\widetilde M_h$ & $\widetilde C_h-\widetilde M_h$\\
\midrule
Terminus-2 & 4.9 & 0.431 & 0.134 & 0.297\\
mini-SWE-agent & 8.6 & 0.631 & 0.335 & 0.296\\
DeepSeek Harness & 8.6 & 0.590 & 0.382 & 0.208\\
Pi coding-agent & 6.6 & 0.383 & 0.206 & 0.177\\
DeepAgents & 14.2 & 0.261 & 0.429 & $-0.168$\\
\bottomrule
\end{tabular}
\end{table}

\paragraph{Action shares.}
Table~\ref{tab:real-agents-actions} gives the mean share of each category at the recorded states of complete target trajectories, for the target and for the teacher sampled at the same states. The remaining categories each hold at most 7.2\% for either model. Across all five harnesses, the target inspects more often than the teacher. It also marks the task complete less often, except under Terminus-2, where the two do so at similar rates.

\begin{table}[htbp]
\centering\small
\caption{Action category shares (\%) at the states of complete target trajectories on the Terminal-Bench Lite probe tasks, target / teacher.}
\label{tab:real-agents-actions}
\setlength{\tabcolsep}{4pt}
\begin{tabular}{@{}lrcccccc@{}}
\toprule
Harness & States & Inspect & Modify & Plan & Stop & Opaque & Invalid\\
\midrule
Terminus-2 & 44 & 37.7 / 30.3 & 15.8 / 20.8 & 0.0 / 0.0 & 38.3 / 35.1 & 8.2 / 6.4 & 0.0 / 0.1\\
mini-SWE-agent & 77 & 59.7 / 43.5 & 14.1 / 11.3 & 0.0 / 0.0 & 11.7 / 22.4 & 14.1 / 18.9 & 0.1 / 0.0\\
DeepSeek Harness & 69 & 33.3 / 20.0 & 12.1 / 15.0 & 15.1 / 26.5 & 11.2 / 14.8 & 27.7 / 18.6 & 0.1 / 2.7\\
Pi coding-agent & 46 & 48.3 / 37.0 & 20.4 / 22.4 & 0.0 / 0.0 & 12.9 / 21.0 & 18.0 / 16.6 & 0.0 / 0.4\\
DeepAgents & 85 & 27.9 / 24.8 & 11.8 / 16.8 & 11.1 / 11.4 & 3.9 / 8.4 & 38.2 / 32.4 & 3.5 / 0.0\\
\bottomrule
\end{tabular}
\end{table}

\subsection{Exact Harness Scores and Task Success}
\label{app:exp-selection}
\paragraph{Tasks and harnesses.}
We evaluate H0--H5 from Table~\ref{tab:e1-harnesses} on the fixed test set of 576 tasks, with 192 tasks from each of the three MiniWoB++ families in Appendix~\ref{app:e1}. We reuse their compiled execution graphs, whose transitions and successful task completion were checked against the native browser. The horizon is ten actor decisions, with remaining budget included in the state. H4 retains the archived decisions of its separate frozen Qwen3.5-2B reviewer.

\paragraph{Frozen model and semantic policy.}
The actor is off-the-shelf Qwen3.5-9B, with fixed weights and no project adapter. Each prompt exposes the harness's complete legal action menu, with a single-token code for every parameterized semantic action. We obtain the action distribution by normalizing the corresponding token logits at temperature one, using BF16 inference, FP64 normalization, and a state microbatch of eight. Thinking is disabled. The objective instruction asks the model to maximize exact task completion within its decision budget. Actor probabilities are computed anew for this model under every harness. We set each harness's reference to this same frozen semantic distribution and hold it fixed when solving for the optimal policy.

\paragraph{Objective and exact computation.}
We use the finite-horizon counterpart of Eq.~\eqref{eq:objective}, with $\gamma=1$, $\tau=0.1$, binary terminal success reward, and zero explicit execution costs. This objective and reference define a new control problem for each harness, so the ceilings differ from those of the post-training study in Section~\ref{sec:exp-training}. Backward induction on the acyclic graph gives $\pistar_\I$ and the initial-state optimal value $C$. Let $\nu^\Pi_\I(s)=\sum_{t=0}^{9}\Pr_\I^\Pi(s_t=s)$ be the expected number of visits to a nonterminal state. Forward propagation under the evaluated policy supplies these visitation weights, giving
\begin{equation}
M(\theta,\I)=\tau\sum_s\nu^{\Pi_{\theta,\I}}_\I(s)
\KL\!\left(\Pi_{\theta,\I}(\cdot\mid s)\middle\Vert\pistar_\I(\cdot\mid s)\right).
\label{eq:finite-success-mismatch}
\end{equation}
The Bellman residual telescopes over the finite horizon to give $C-J=M$. Since the evaluated policy equals its fixed reference, its regularization penalty is zero, and $J$ equals task success probability. We compute $J$ by forward reward accumulation and success by a separate backward recursion. These independent computations agree with $C-M$ to numerical precision. The largest observed identity residual is $7.22\times10^{-16}$.

\begin{figure}[t]
\centering
\includegraphics[width=0.8\linewidth]{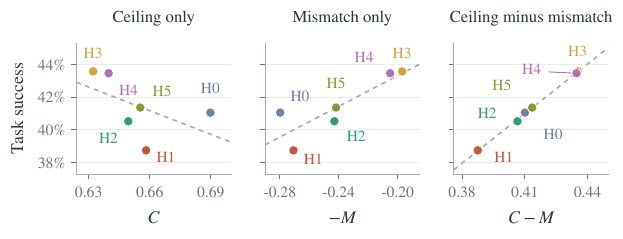}
\caption{Harness scores compared with task success. Each point is one harness evaluated with the same frozen Qwen3.5-9B on a shared task set. The panels compare $C$, $-M$, and $C-M$ with rollout success. Dashed lines show linear fits.}
\label{fig:exact-harness-scores}
\end{figure}

\paragraph{Rollout evaluation and aggregation.}
For each task and harness, we draw 256 independent trajectories from the measured actor distribution on the compiled graph. The fraction that reaches native task success supplies the vertical coordinate in Figure~\ref{fig:exact-harness-scores}.

 Scores and success are averaged within each family and then equally across families. Each point therefore compares the same task distribution. The three selection rules maximize $C$, $-M$, and $C-M$, respectively. The dashed lines are unweighted linear fits through the six displayed harness means. All actor evaluations run on four RTX 4090 GPUs.

For these deterministic task graphs with binary terminal reward, an individual task's reference success probability $S$ gives $C=\tau\log(1+S(e^{1/\tau}-1))$. This is increasing in $S$, but averaging this nonlinear quantity across tasks can order harnesses differently from averaging success. In the aggregate comparison, ceiling selects H0, while both negative mismatch and $C-M$ select H3. Negative mismatch ranks H1 above H0, whereas exact and sampled success rank H0 above H1. Combining ceiling and mismatch gives the complete ordering of the exact mean success probabilities.

\end{document}